\documentclass[letterpaper]{article} % DO NOT CHANGE THIS
\usepackage{aaai2026}  % DO NOT CHANGE THIS
\nocopyright
\usepackage{times}  % DO NOT CHANGE THIS
\usepackage{helvet}  % DO NOT CHANGE THIS
\usepackage{courier}  % DO NOT CHANGE THIS
\usepackage[hyphens]{url}  % DO NOT CHANGE THIS
\usepackage{graphicx} % DO NOT CHANGE THIS
\usepackage{natbib}  % DO NOT CHANGE THIS AND DO NOT ADD ANY OPTIONS TO IT
\usepackage{caption} % DO NOT CHANGE THIS AND DO NOT ADD ANY OPTIONS TO IT
\usepackage{algorithm}

\usepackage{newfloat}
\usepackage{listings}

\usepackage{pifont}

\usepackage{amsmath}
\usepackage{array}
\usepackage{graphicx}
\usepackage{cite}

\usepackage{booktabs}
\usepackage{multirow}
\usepackage{subcaption}

\usepackage{colortbl}

 \usepackage{multirow}

\usepackage{algpseudocode}

\usepackage[utf8]{inputenc} % allow utf-8 input
\usepackage[T1]{fontenc}    % use 8-bit T1 fonts
\usepackage{booktabs}       % professional-quality tables
\usepackage{amsfonts}       % blackboard math symbols
\usepackage{nicefrac}       % compact symbols for 1/2, etc.
\usepackage{microtype}      % microtypography
\usepackage{algpseudocode} % 提供 algorithmic 伪代码环境
\usepackage{xpatch}
\makeatletter
\xapptocmd{\NAT@bibsetnum}{\setlength{\leftmargin}{0pt}\setlength{\itemindent}{\labelwidth}\addtolength{\itemindent}{\labelsep}}{}{}

\DeclareCaptionStyle{ruled}{labelfont=normalfont,labelsep=colon,strut=off} % DO NOT CHANGE THIS
\floatstyle{ruled}
\newfloat{listing}{tb}{lst}{}
\floatname{listing}{Listing}
\newenvironment{breakablealgorithm}
  {% \begin{breakablealgorithm}
   \begin{center}
     \refstepcounter{algorithm}% New algorithm
     \hrule height.8pt depth0pt \kern2pt% \@fs@pre for \@fs@ruled
     \renewcommand{\caption}[2][\relax]{% Make a new \caption
       {\raggedright\textbf{\ALG@name~\thealgorithm} ##2\par}%
       \ifx\relax##1\relax % #1 is \relax
         \addcontentsline{loa}{algorithm}{\protect\numberline{\thealgorithm}##2}%
       \else % #1 is not \relax
         \addcontentsline{loa}{algorithm}{\protect\numberline{\thealgorithm}##1}%
       \fi
       \kern2pt\hrule\kern2pt
     }
  }{% \end{breakablealgorithm}
     \kern2pt\hrule\relax% \@fs@post for \@fs@ruled
   \end{center}
  }

\title{KVmix: Gradient-Based Layer Importance-Aware Mixed-Precision Quantization for KV Cache}
\author{
    Fei Li, %%\textsuperscript{\rm 1}
    Song Liu\thanks{Corresponding author.},
    Weiguo Wu,
    Shiqiang Nie,
    Jinyu Wang
}
\affiliations{
    School of Computer Science and Technology, Xi’an Jiaotong University\\
    lifei@stu.xjtu.edu.cn, \{liusong, wgwu, shiqiang.nie, jinyu.wang\}@xjtu.edu.cn

}

\usepackage{bibentry}
\begin{document}

\maketitle

\begin{abstract}
The high memory demands of the Key-Value (KV) Cache during the inference of Large Language Models (LLMs) severely restrict their deployment in resource-constrained platforms. Quantization can effectively alleviate the memory pressure caused by KV Cache. However, existing methods either rely on static one-size-fits-all precision allocation or fail to dynamically prioritize critical KV in long-context tasks, forcing memory-accuracy-throughput tradeoffs. In this work, we propose a novel mixed-precision quantization method for KV Cache named KVmix. KVmix leverages gradient-based importance analysis to evaluate how individual Key and Value projection matrices affect the model loss, enabling layer-specific bit-width allocation for mix-precision quantization. It dynamically prioritizes higher precision for important layers while aggressively quantizing less influential ones, achieving a tunable balance between accuracy and efficiency. KVmix introduces a dynamic long-context optimization strategy that adaptively keeps full-precision KV pairs for recent pivotal tokens and compresses older ones, achieving high-quality sequence generation with low memory usage. Additionally, KVmix provides efficient low-bit quantization and CUDA kernels to optimize computational overhead. On LLMs such as Llama and Mistral, KVmix achieves near-lossless inference performance with extremely low quantization configuration (Key 2.19bit Value 2.38bit), while delivering a remarkable 4.9× memory compression and a 5.3× speedup in inference throughput. %%The source code of this paper is included in the supplementary materials and will be open-sourced in a later phase. %%Code is available at https://anonymous.4open.science/r/KVmix-1386/.

\end{abstract}

\begin{links}
    \link{Code}{https://github.com/LfLab-AI/KVmix}
    % \link{Datasets}{https://aaai.org/example/datasets}
    % \link{Extended version}{https://aaai.org/example/extended-version}
\end{links}

\section{Introduction}
Large Language Models (LLMs) \cite{vaswani2017attention}, such as GPT \cite{radford2019language}, Llama \cite{touvron2023llama1}, and their derivatives, have significantly advanced the field of Natural Language Processing (NLP). These models exhibit outstanding performance \cite{hadi2023survey,chang2024survey} across a diverse array of tasks, including text generation, question answering, and machine translation. The Key-Value (KV) Cache plays an essential role in the autoregressive decoding process of LLMs. The KV Cache substantially reduces redundant computations in the attention mechanism by storing KV states from preceding time steps for subsequent token generation \cite{xiao2023efficient}. Nevertheless, as sequence lengths grow, the memory footprint of the KV Cache increases linearly, presenting a formidable challenge to hardware resources. For instance, a 70B-parameters model may require over 50GB of memory to maintain the KV Cache for a 20k-token sequence, exceeding typical GPU capacity. In scenarios involving multiple concurrent requests, the KV Cache for each request cannot be shared due to its dependence on unique preceding prompts. Although model parameters can be reused, memory quickly becomes saturated due to the KV Cache demands. Once memory is depleted, data is offloaded to system memory or even disks, resulting in frequent High Bandwidth Memory (HBM) exchanges with system memory. This process causes latency to surge exponentially, leading to catastrophic performance degradation.

The KV Cache's characteristics outlined above severely restrict LLM deployment and inference efficiency in resource-constrained environments, underscoring the pressing need for efficient memory optimization \cite{liu2024cachegen}. Recent research tackling this issue has predominantly focused on reducing the memory overhead of the KV Cache through quantization and sparsification techniques \cite{shi2024keep,adnan2024keyformer}. Quantization methods, in particular, have gained widespread adoption in industry, significantly contributing to the scalability and accessibility of large-scale models \cite{kumar2024residual}. Quantizing the KV Cache can markedly reduce memory usage. Existing quantization methods have demonstrated impressive model performance even at very low bit-widths. However, these methods either rely on static one-size-fits-all precision allocation schemes \cite{liu2024kivi,liu2024intactkv}, lacking flexibility and performance-aware adaptation capabilities, or incur high computational costs of dynamic quantization while failing to adaptively prioritize critical KVs in long-context tasks \cite{dong2024qaq,duanmu2024skvq}. Therefore, they are forced to make suboptimal trade-offs among memory usage, model accuracy, and computational throughput.

%%—for instance, transitioning from FP16 to INT8 halves memory requirements while accelerating memory access and improving throughput.
%%,liu2024unlocking

To address these problems, this paper proposes KVmix, a novel mixed-precision quantization method for KV Cache. Compared to existing mixed quantization methods \cite{dong2024qaq,li2025kvtuner}, KVmix analyzes the importance differences of different model layers at a very low cost, thereby allowing for flexible modification of the quantization configuration based on the model's performance requirements. This flexibility enables KVmix to maximize the compression rate of the KV Cache and the throughput of the model while maintaining controllable precision. Our specific contributions are as follows:

\begin{itemize}
\item We propose a novel layer importance-aware mixed-precision quantization method. This method assesses the importance of KVs at each layer by computing the $L2$ gradient norms of Key and Value projection weights with respect to the model’s loss function. Based thereon, it independently applies mixed-precision quantization to different layers, allocating higher bit-widths to critical layers and lower to less influential ones. Therefore, it provides the flexibility to balance between accuracy and resource efficiency across diverse inference scenarios.
%%As a result, KVmix achieves dynamic optimization of model accuracy and resource utilization.
\item We propose a dynamic pivotal context selection strategy to optimize long-context tasks. According to the KV importance analysis, it adaptively keeps full-precision KV pairs for recent pivotal tokens while aggressively compressing older pairs. This strategy ensures high-quality sequence generation in long-context inference scenarios while dynamically reducing the number of full-precision KV pairs for better memory optimization.
\item We design efficient CUDA implementations and a high-compression 3-bit quantization method for KVmix. Extensive experimental results show that KVmix achieves nearly lossless model accuracy across multiple LLMs and datasets, with a 4.9$\times$ memory usage reduction and a 5.3$\times$ speedup in inference efficiency, outperforming prior state-of-the-art (SOTA) quantization methods for KV Cache.
\end{itemize}

\section{Related Work and Motivation}
\subsection{Related Work}

To mitigate KV Cache memory challenges, researchers have developed many optimization methods, primarily centered on compression techniques and dynamic memory management \cite{kwon2023efficient,lee2024infinigen}. We mainly discuss the KV Cache compression techniques related to this work. Existing compression approaches encompass quantization, sparsification \cite{zhang2023h2o,li2024snapkv}, and KV Cache sharing \cite{sun2024you, wu2024layer}. Our method is orthogonal to existing weight quantization \cite{frantar2022gptq, lin2023awq} and sparsification methods, and it can also be used as a guideline for the importance of different layers during KV sparsification to achieve more accurate KV eviction.

Extensive research has focused on reducing the memory overhead of KV Cache through quantization. For example, KIVI \cite{liu2024kivi} introduced a 2-bit asymmetric quantization technique, employing per-channel quantization for Keys and per-token quantization for Values. KVQuant \cite{hooper2024kvquant} proposed a non-uniform quantization strategy, integrating pre-RoPE per-channel Key quantization with per-token Value quantization. It employs offline calibration to manage outliers, achieving robust performance in long-context inference scenarios. QAQ \cite{dong2024qaq} developed a dynamic mixed-precision quantization method that calculates quantization bits for Keys and Values online and optimizes the trade-off between accuracy and compression ratio by predicting attention scores. Atom \cite{zhao2024atom} investigated a mixed-precision scheme involving 4-bit and 8-bit activations, dynamically quantizing activations to adapt to input distributions. QJL \cite{zandieh2025qjl} introduced a 1-bit quantization technique for the Key, leveraging the Johnson-Lindenstrauss transform followed by sign-bit quantization. KVTuner \cite{li2025kvtuner} frames the mixed quantization of KV caches as a search optimization problem, aiming to find the optimal KV quantization configuration within a vast search space. Numerous other studies have also made significant contributions to KV Cache quantization \cite{yue2024wkvquant,yang2024no,liu2024intactkv}. These efforts generally aim to quantize as many KVs as possible to the lowest feasible bit-width. While methods such as KVQuant, QAQ, and KVTuner have introduced mixed-precision quantization for KV,  they often entail significant computational or search overhead. In contrast, this work introduces a lightweight and flexible framework that allows users to dynamically balance quantization bit-width and model accuracy based on specific deployment requirements.

Moreover, several works have identified the attention sink phenomenon, where attention scores excessively favor initial or recent tokens, and newly generated tokens emphasize recent contexts. StreamingLLM \cite{xiao2023efficient} leverages this characteristic to achieve efficient infinite-length streaming inference by retaining attention sinks (e.g., initial tokens) alongside recent tokens. PyramidInfer \cite{yang2024pyramidinfer} applies layer-wise KV cache compression, selectively keeping key contexts based on attention patterns. Inspired by these works, we propose a dynamic pivotal context selection strategy. Unlike prior approaches, ours determines pivotal context size based on layer-specific KV importance analysis and dynamically updates full-precision KV pairs during decoding, better balancing memory efficiency and generation quality.

\subsection{Motivation} \label{Motivation}
\begin{figure}[htbp]
	\centering
	\includegraphics[width=1.0\linewidth]{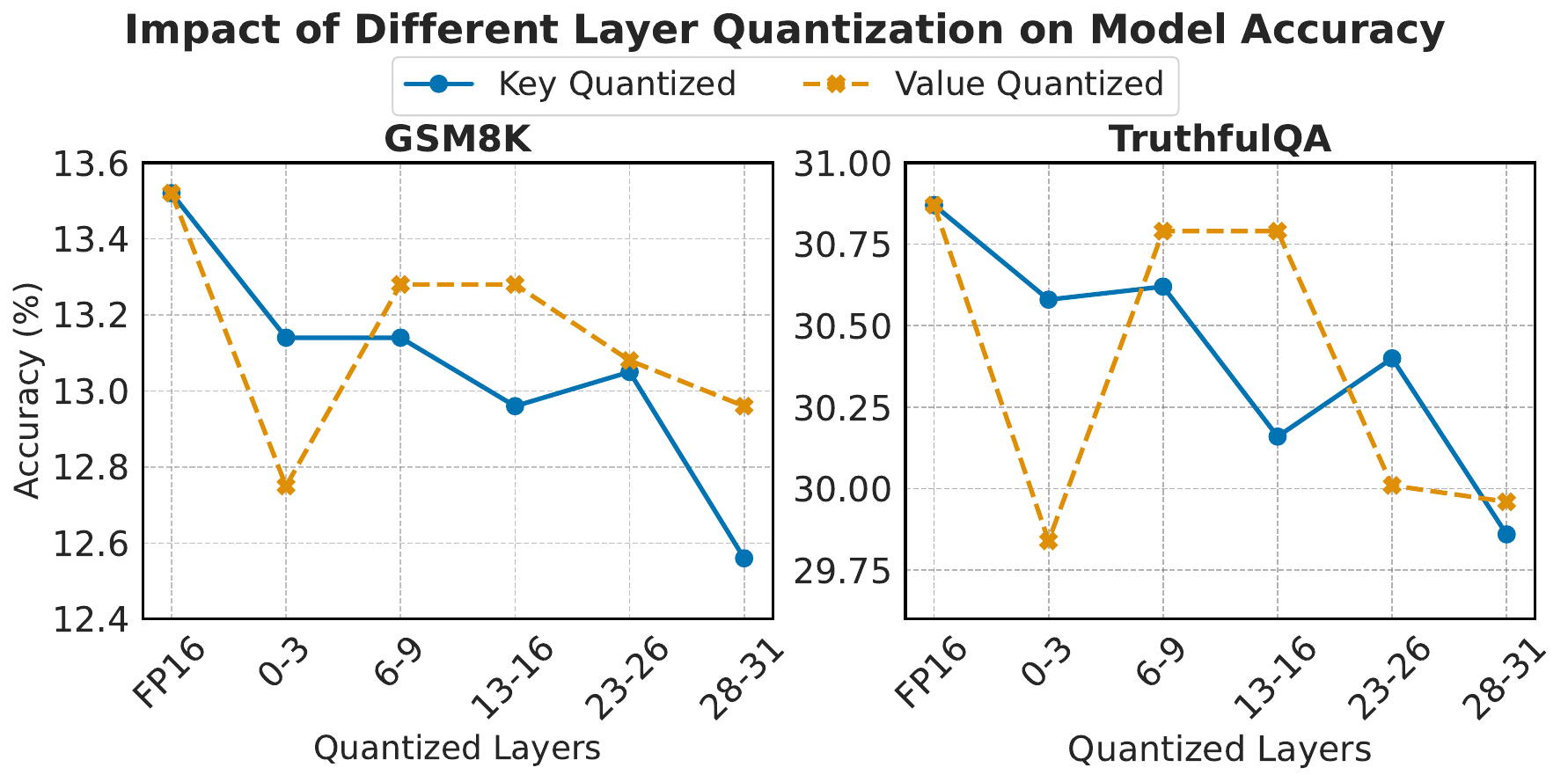}
	\caption{Accuracy of the Llama 2-7B model \cite{touvron2023llama} on the GSM8K \cite{cobbe2021training} and TruthfulQA \cite{lin-etal-2022-truthfulqa} datasets using lm\_eval \cite{eval-harness} (FP16 represents no quantization; 0-3 indicates 2-bit quantization applied individually to the Key or Value of layers 0 through 3, respectively. And so on).}
	\label{fig01}
\end{figure}

Current KV Cache quantization methods rely on fixed quantization strategies, neglecting the varying contributions of KV across layers to the final output. To substantiate that quantizing Keys or Values from different layers impacts the model differently, we selectively applied 2-bit quantization to the Keys or Values of distinct layers and evaluated the effects on the model’s accuracy, with results presented in Fig. \ref{fig01}. The findings demonstrate that quantizing Keys or Values from different layers has varying impacts on the model’s generation quality. However, \textit{efficiently analyzing the contribution disparities across layers of the model to allocate different quantization bits to Keys or Values remains a critical challenge that needs to be addressed.}

\begin{figure}[htbp]
	\centering
	\includegraphics[width=1.0\linewidth]{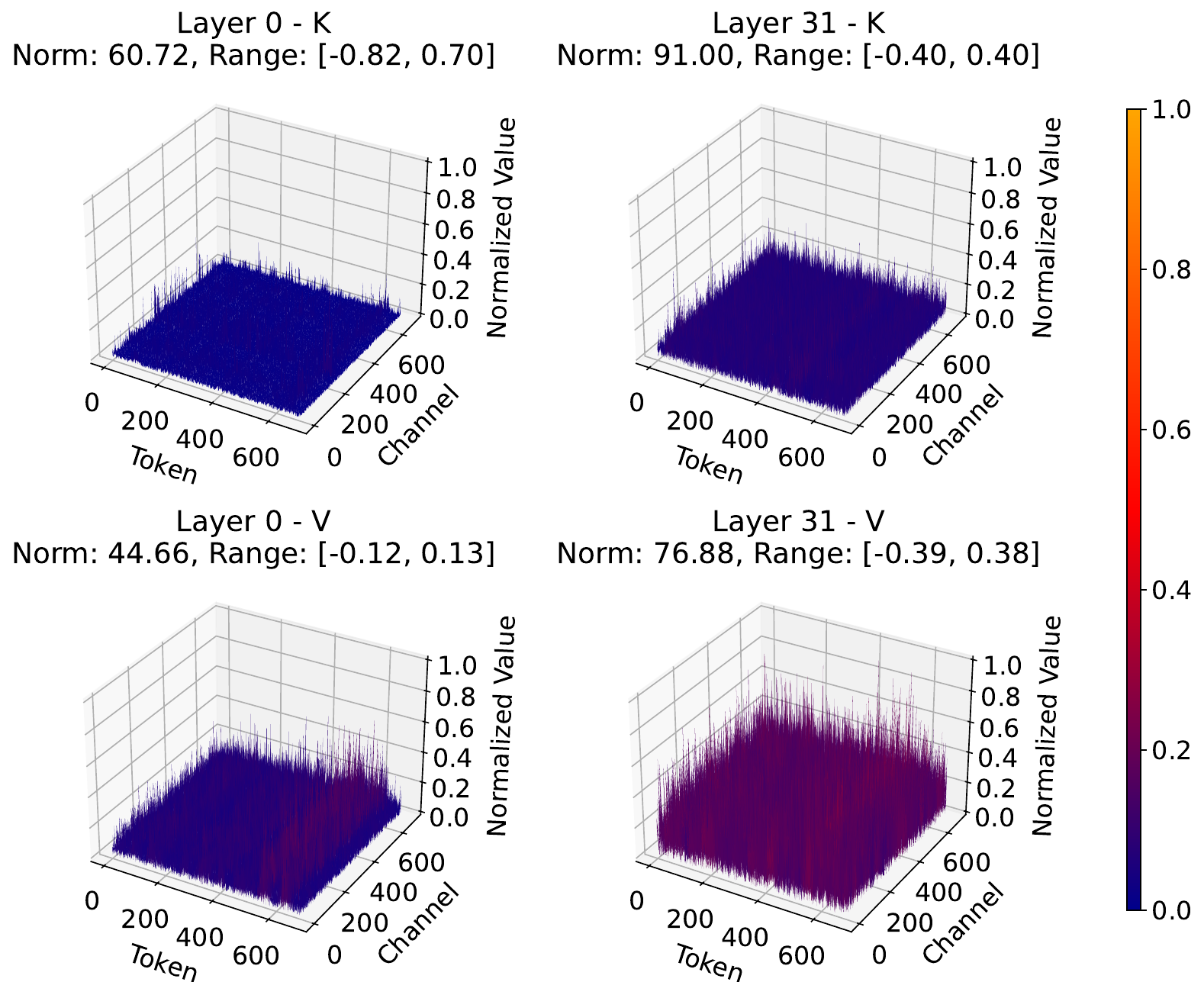}
	\caption{Projection matrix weights of K and V across different layers for the Llama 2-7B model. "Norm" represents the L2 norm of weight matrix for each layer, while "Range" indicates the range of values within each layer's weight matrix.}
	\label{fig02}
\end{figure}

In each layer of the KV Cache, the computation process works as follows: at time step $t$, the $i$-th layer receives hidden states $H_{i-1,t}$ from the preceding layer's output. These hidden states are used to compute the current token's K and V as: $ K_{i,t} = W_{k_i} \cdot H_{i-1,t} $ and $ V_{i,t} = W_{v_i} \cdot H_{i-1,t} $, where $ W_{k_i} $ and $ W_{v_i} $ are the projection weights for K and V at the $ i $-th layer. After computation, the computed $ K_{i,t} $ and $ V_{i,t} $ are concatenated with the previously stored KV, yielding the complete K and V sequences up to time step $ t $: $ K_{i,1:t} = [K_{i,1}, K_{i,2}, \dots, K_{i,t}] $ and $ V_{i,1:t} = [V_{i,1}, V_{i,2}, \dots, V_{i,t}] $. This computation process indicates that the $ W_{k_i} $ and $ W_{v_i} $ determine how Key and Value are extracted from the hidden states, directly affecting the quality of the KV pairs generated by the attention mechanism and the layer's contribution to the model's output. Fig. \ref{fig02} provides heatmaps of the $W_k$ and $W_v$ for the Llama 2-7B model, with more comprehensive heatmaps for additional models included in Technical Appendix A. The heatmaps highlight two key insights: \ding{172} \textit{Significant variations in KV weight values across different layers.} \ding{173} \textit{Distinct distribution patterns of KV weights within the same layer.} For KVs, their values dynamically adapt to changes in the input; however, $W_k$ and $W_v$ are learned during the model’s training phase and remain static during inference. As a result, these weights can be leveraged to assess the importance differences of Keys and Values across layers.

%%, with more comprehensive heatmaps for additional models included in Technical Appendix A

\section{Methodology}
\subsection{KV Importance Analysis} \label{KVmix Profiler}

In the attention mechanism, KV is computed by applying the KV projection weight matrices to the hidden state of the previous layer for linear transformation, then combining the result with the Query vector to calculate the attention output. Therefore, the magnitude of the KV projection weights alone is insufficient to measure the importance of KV across layers, necessitating a more precise evaluation metric. This metric should quantify the sensitivity of K and V at each layer to the model's loss function $L$. Based on the Motivation and by using the chain rule, we can obtain:
\begin{align}
&K_{i,t} = W_{k_i} \cdot H_{i-1,t} \\ %%\notag
&\longrightarrow \nabla_{W_{k_i}} L = \frac{\partial L}{\partial K_i} \frac{\partial K_i}{\partial W_{k_i}} = \frac{\partial L}{\partial K_i} H_{i-1}^T \\
&\longrightarrow || \frac{\partial L}{\partial K_i} ||_2 = \frac{|| \nabla_{W_{k_i}} L ||_2}{|| H_{i-1} ||_2}
\end{align}
\begin{align}
&V_{i,t} = W_{v_i} \cdot H_{i-1,t} \\
&\longrightarrow \nabla_{W_{v_i}} L = \frac{\partial L}{\partial V_i} \frac{\partial V_i}{\partial W_{v_i}} = \frac{\partial L}{\partial V_i} H_{i-1}^T \\
&\longrightarrow || \frac{\partial L}{\partial V_i} ||_2 = \frac{|| \nabla_{W_{v_i}} L ||_2}{|| H_{i-1} ||_2}
\end{align}
Here, $\nabla_{W_{k_i}} L$ and $\nabla_{W_{v_i}} L$ denote the gradients of $L$ with respect to $W_{k_i}$ and $W_{v_i}$, respectively. $\frac{\partial L}{\partial K_i}$ and $\frac{\partial L}{\partial V_i}$ represent the partial derivatives of $L$ with respect to the Key and Value, respectively. To quantify the perturbation, assume that the quantization operation introduces small perturbations to the Key and Value matrices, i.e., $K_i^q = K_i + \Delta K$ and $V_i^q = V_i + \Delta V$, where $\Delta K$ and $\Delta V$ are the quantization errors. The change in $L$ due to quantization is $\Delta L = L(K^q, V^q) - L(K, V)$, which is the difference between the original loss and the quantized loss. To approximate $\Delta L$, perform a first-order Taylor expansion around Key:
\begin{align}
L(K^q, V^q) \approx L(K, V) + \frac{\partial L}{\partial K} \cdot \Delta K + \frac{\partial L}{\partial V} \cdot \Delta V
\end{align}
Thus, the loss change due to quantization is:
\begin{align}
\Delta L \approx \frac{\partial L}{\partial K} \cdot \Delta K + \frac{\partial L}{\partial V} \cdot \Delta V.
\end{align}
$\frac{\partial L}{\partial K}$ and $\frac{\partial L}{\partial V}$ are the gradients $\frac{\partial L}{\partial K_i}$ and $\frac{\partial L}{\partial V_i}$ computed earlier, so: 
\begin{align}
\Delta L \approx \langle \frac{\partial L}{\partial K_i}, \Delta K \rangle + \langle \frac{\partial L}{\partial V_i}, \Delta V \rangle
\end{align}
where $\langle \cdot, \cdot \rangle$ denotes the inner product.
For a fixed $\Delta K$, a larger $\left\|\frac{\partial L}{\partial K_i}\right\|_2$ (take L2 norm) amplifies $\Delta L$, indicating greater sensitivity. The weight gradient norm thus reflects Key's impact on $L$, as $\|\nabla_{W_{k_i}} L\|_2$ proxies $\left\|\frac{\partial L}{\partial K_i}\right\|_2$ (modulo input scaling). The Value is the same.

Based on the above analysis, we propose the KVmix profiler, a gradient-based method that quantifies the contribution of each layer’s Key and Value to the model’s output, enabling a judicious mixed-precision quantization strategy. Specifically, we compute the $L2$ norm of the gradients of the model’s loss function $L$ with respect to the Key and Value projection weight matrices for each model layer ($\|\nabla_{W_{k_i}} L\|_2$ and $\|\nabla_{W_{v_i}} L\|_2$), and evaluate the importance of the Key and Value components based on these $L2$ norm values. KVmix profiler captures the dynamic sensitivity of these parameters during the model inference process, and provides a layer-specific importance metric to support efficient mixed-precision quantization in subsequent inference stages.

The implementation of KVmix profiler consists of the following three key steps: \textbf{\ding{172} Data preparation and forward propagation}. A full-precision model is loaded, and multiple prompts are randomly sampled from a target dataset to serve as input data. These prompts are tokenized into input tensors using the tokenizer. Leveraging the autoregressive property of LLMs, each input tensor is shifted left by one position to be used as the corresponding label tensor for computing the model’s loss function. Subsequently, the loss value for each input is determined through forward propagation. \textbf{\ding{173} Gradient calculation and importance assessment}. For the $i$-layer of the model, the gradients of the losses with respect to the projection weights of the Key ($W_{k_i}$) and Value ($W_{v_i}$) are computed independently. This process begins with backpropagation to calculate the gradients, i.e., $\nabla_{W_{k_i}} L$ and $\nabla_{W_{v_i}} L$. The magnitude of these gradients is then evaluated using the $L2$ norm, i.e., $\| \nabla_{W_{k_i}} L \|_2$  and $\| \nabla_{W_{v_i}} L \|_2$. The importance scores of the Keys and Values for each layer can be expressed as:
\begin{equation}
s_{k_i} = \| \nabla_{W_{k_i}} L \|_2, 
s_{v_i} = \| \nabla_{W_{v_i}} L \|_2
\end{equation}
A larger $s_{k_i}$ or $s_{v_i}$ signifies a greater impact of that $i$-th layer’s Key or Value on model’s output. To enhance assessment reliability, the gradient norms can be averaged across multiple prompts ($p$), yielding an average importance score for each layer's Key and Value ($P$ is the number of prompts): 
\begin{equation}
\bar{s}_{k_i} = \frac{1}{P} \sum_{p=1}^{P} s_{k_i}^{(p)}, 
\bar{s}_{v_i} = \frac{1}{P} \sum_{p=1}^{P} s_{v_i}^{(p)}
\end{equation}
We classify the importance of the Key and Value components across all model layers using the importance scores. The top 20\% of the layers of $\bar{s}_{k_i}$ and $\bar{s}_{v_i}$ are quantized with high-bit representations (e.g., 3-bit or 4-bit), while the remaining 80\% of layers adopt more aggressive low-bit quantization (e.g., 2-bit). This 20\%-80\% split is not fixed and can be dynamically adjusted according to the requirements to balance the trade-off between model accuracy and memory usage. Increasing the proportion of low-bit quantization layers can further reduce the memory usage of the KV Cache, but may sacrifice some accuracy. \textbf{\ding{174}Model Configuration and Inference}. The KV quantization configuration results derived from the above steps are incorporated into the model configuration, enabling the quantized model to be used for inference. The detailed workflow is depicted in Fig. \ref{fig03}, with the algorithmic procedure outlined in Algorithm 1 of Appendix B. The profiling is performed offline and therefore does not affect inference efficiency. Moreover, the profiling is performed once, allowing the model to reuse the results for subsequent inference tasks.
%%, with the algorithmic procedure outlined in Algorithm 1 of Appendix B

\begin{figure*}[htbp]
	\centering
	\includegraphics[width=1.0\linewidth]{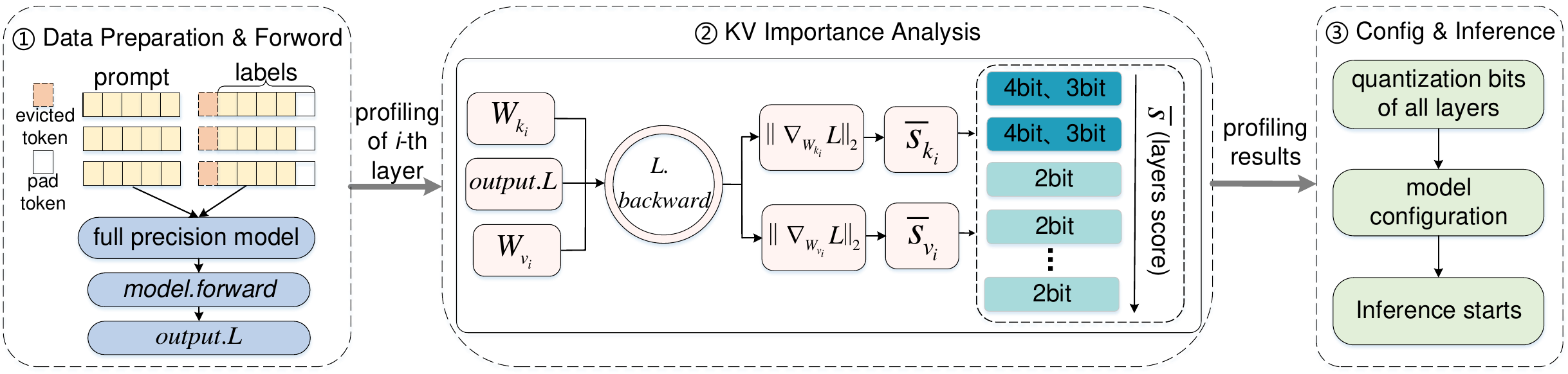}
	\caption{The overview of KVmix profiler.}
	\label{fig03}
\end{figure*}

\subsection{Asymmetric Low-Bit Quantization}

\subsubsection{Asymmetric Quantization Strategy}

We use per-channel and per-token grouping quantization methods for Key and Value, respectively. The KV Cache has the shape $[B, nh, T, D]$, where $B$ is the batch size, $nh$ is the number of attention heads, $T$ is the token sequence length, and $D$ is the head dimension. When the Key is quantized per channel ($D$), the tensor is reshaped to $[B \times nh \times D, T]$, with each group comprising all tokens of a single channel. This approach is inspired by the distributional properties of the Key Cache that exhibit significant outliers in the channel dimension, i.e., certain channels exhibit significantly large magnitude values. Per-channel quantization isolates errors within each channel and prevents outliers from affecting other channels. When Value is quantized per token, the shape of the tensor is preserved, and each group contains all channels of a single token. Unlike the Key Cache, the Value Cache has no pronounced outliers, but plays a critical role in computing the attention output. Per-token quantization confines errors to individual tokens, preserving the integrity of other important tokens. This asymmetric quantization strategy effectively reduces errors introduced during KV Cache quantization.

\subsubsection{Group-Wise Low-Bit Quantization}
We use group-wise low-bit quantization to minimize KV Cache memory usage. The process includes: \ding{172} Calculation of scaling factor $s$. For each group (per-channel for Key, per-token for Value), compute $s = \frac{max\_val - min\_val}{q_{max}}$ using group min/max values, where $q_{max}$ denotes the maximum quantized value. \ding{173} Element quantization. Quantize elements with $ q = round \left( \frac{x - min\_val}{s} \right)$, where $x$ represents the original element value, and $q$ is the quantized value. \ding{174} Clipping. Limit $q$ to $\max(0, \min(q, {q_{max}}))$. \ding{175} Storage and dequantization. The quantized values are stored using bit operations within a 32-bit integer ($int32$). For 4, 2, and 1 bit, the number of elements per $int32$ is: $feat\_per\_int=32/bit$. Dequantization is performed by $x = q \cdot s + min\_val$. For \textbf{3-bit quantization}, we introduce a new packing strategy to maximize memory efficiency. We organize the quantized elements into blocks of 11, each stored in a 32-bit integer, with the first 10 elements quantized to 3 bits and the 11th element to 2 bits. The clipping range is adjusted based on the element index:
\begin{equation}
q_{max} = 
\begin{cases} 
7, & i = 0,1,...,9 \\
3, & i = 10 
\end{cases}
\end{equation}
where $i$ is the element index within a block. This strategy increases packing density by 10\% over uniform 3-bit quantization that can only hold 10 elements per $int32$.

\subsection{Dynamic Pivotal Context Selection}

\begin{figure}[htbp]
	\centering
	\includegraphics[width=0.9\columnwidth]{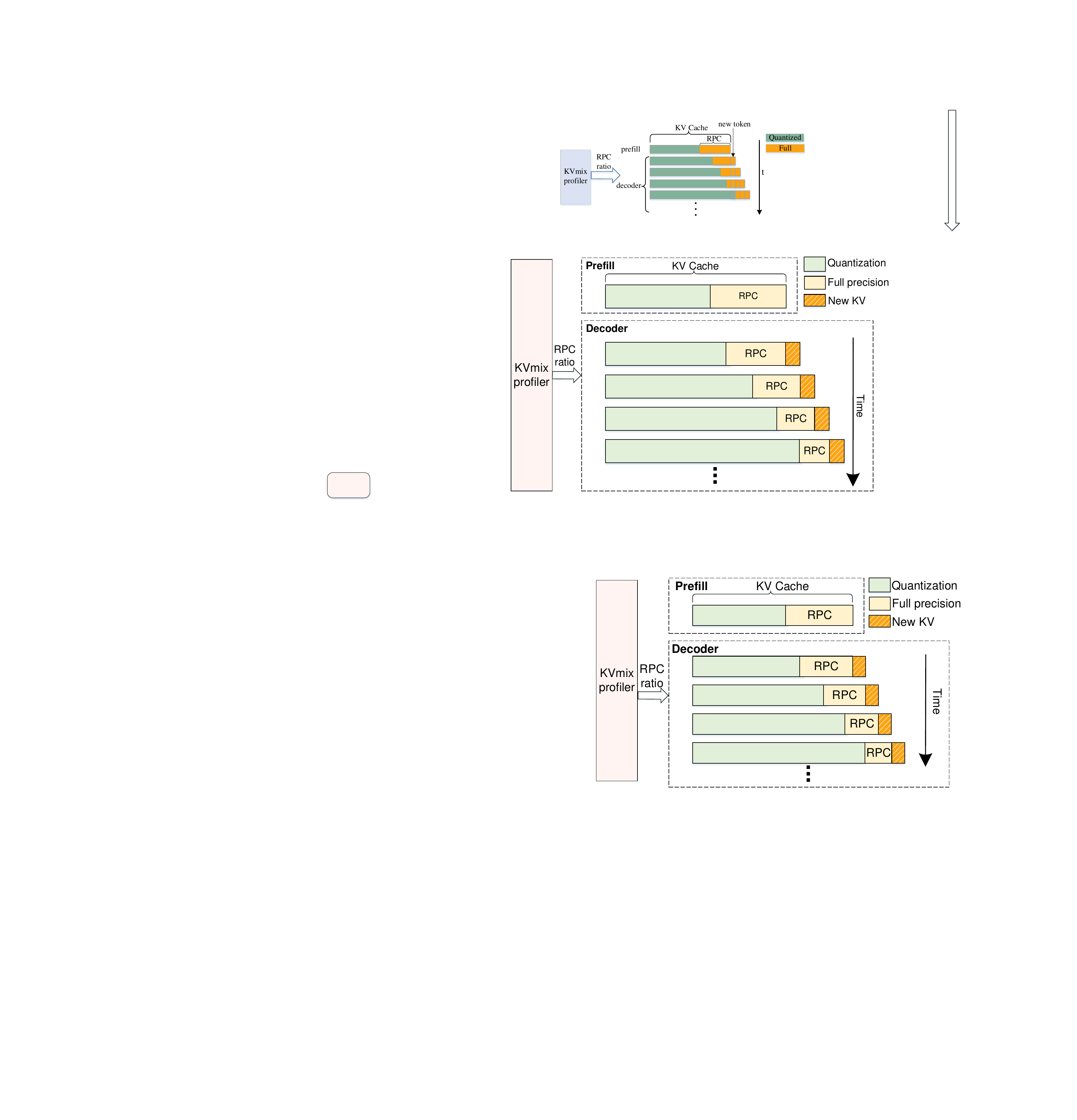}
	\caption{Dynamic adjustment of quantized KV Cache based on RPC during prefill and decoding phases.}
	\label{fig04}
\end{figure}

In the KV Cache, not all Keys and Values are equally important for generating future tokens. Recent tokens provide the most relevant contextual information for the generation of subsequent tokens and typically have more impact on the tokens to be generated. We define the KVs corresponding to these pivotal recent tokens as the Recent Pivotal Context (RPC). To optimize model performance while minimizing memory usage, we propose a dynamic RPC selection strategy based on the importance analysis provided by KVmix profiler. Specifically, for the $i$-layer, we assign it an RPC selection ratio $r$ based on the $\bar{s}_{k_i}$ and $\bar{s}_{v_i}$ scores, with higher $\bar{s}_{k_i}$ and $\bar{s}_{v_i}$ resulting in larger $r$. The number of RPCs is computed by $num\_RPC = \lfloor r \times current\_RPC \rfloor$. $current\_RPC$ is the sum of the number of new KV states at the current time step and the number of historical RPCs. The corresponding number of KV pairs is selected as RPCs based on $num\_RPC$. We keep full precision for RPCs while performing mixed quantization for less critical and older KV pairs, illustrated in Fig. \ref{fig04}. This strategy ensures that the number of full-precision RPCs is dynamically reduced in runtime during long context inference, thus avoiding excessive memory pressure caused by preserving a large number of full-precision KV pairs, while maintaining high-quality sequence generation. Additionally, since the importance of Key and Value may differ within the same layer, the RPC selection ratio for Key and Value varies accordingly within that layer. The RPC selection ratio can be adjusted to balance accuracy and memory: increasing it improves accuracy but requires more memory. 
%%The RPC selection ratio can be adjusted to provide flexibility in the dynamic balance between inference accuracy and memory usage.

\subsection{ CUDA Implementation}

During model inference, the quantization of the KV Cache introduces additional overhead due to quantization and dequantization operations. To improve inference efficiency, we design efficient CUDA kernels for quantization, dequantization, and matrix-vector multiplication. \textbf{\ding{172} Fusion of quantization and concatenation.} In the decoding phase, the KV states of current layer are concatenated with the historical KV Cache. Quantizing the current states before concatenation causes extra memory access overhead. We fuse quantization and concatenation into a single CUDA kernel, processing each element in a streaming manner. KV elements are quantized and appended directly to the historical KV Cache, thereby reducing memory access. CUDA thread blocks process tokens in parallel, and shared memory caches intermediate results to enhance data locality. \textbf{\ding{173} Fusion of dequantization and matrix-vector multiplication.} In attention computation, the quantized KV requires dequantization before matrix-vector multiplication. Dequantizing the full KV beforehand increases memory usage. We fuse dequantization with multiplication, dequantizing each element on-the-fly and immediately multiplying and accumulating it with its corresponding element, minimizing memory overhead. \textbf{\ding{174} Efficient kernels for multi-bit quantization configurations.} To support KVmix’s various quantization bit-widths, we develop CUDA kernels for 1-, 2-, 3-, and 4-bit quantization, along with tailored matrix-vector multiplication kernels for each configuration, ensuring compatibility across bit-widths.

\section{Experimental Results}
\subsection{Experimental Setup} \label{Experimental Setup}

We evaluated the proposed method using Llama 2-7B-hf, Llama 3-8B-Instruct, Llama 3.1-8B \cite{grattafiori2024llama}, Mistral-7B-Instruct-v0.3 \cite{jiang2023mistral7b}, and Falcon-7B \cite{almazrouei2023falcon} models. The datasets were selected based on three distinct evaluation schemes: \ding{172} Long Context Evaluation: We used the LongBench \cite{bai2024longbench} benchmark to assess performance on long-context tasks. It encompasses multiple key long-text application scenarios. Due to the limited GPU memory, the maximum sequence length was set to 4096. \ding{173} Language Modeling: We measured the perplexity on the Wikitext-2 \cite{merity2016pointer} dataset to evaluate its language modeling capabilities. \ding{174} Mathematical Reasoning: We employed the GSM8K \cite{cobbe2021training} dataset to assess the model's performance on mathematical reasoning tasks. We used the NVIDIA RTX 4090 GPU (24GB) to evaluate the model's inference efficiency and the KV cache's compression rate. 
\subsection{Profiling Results} \label{Profiling Results}
\begin{figure}[htbp]
	\centering
	\includegraphics[width=1.0\columnwidth]{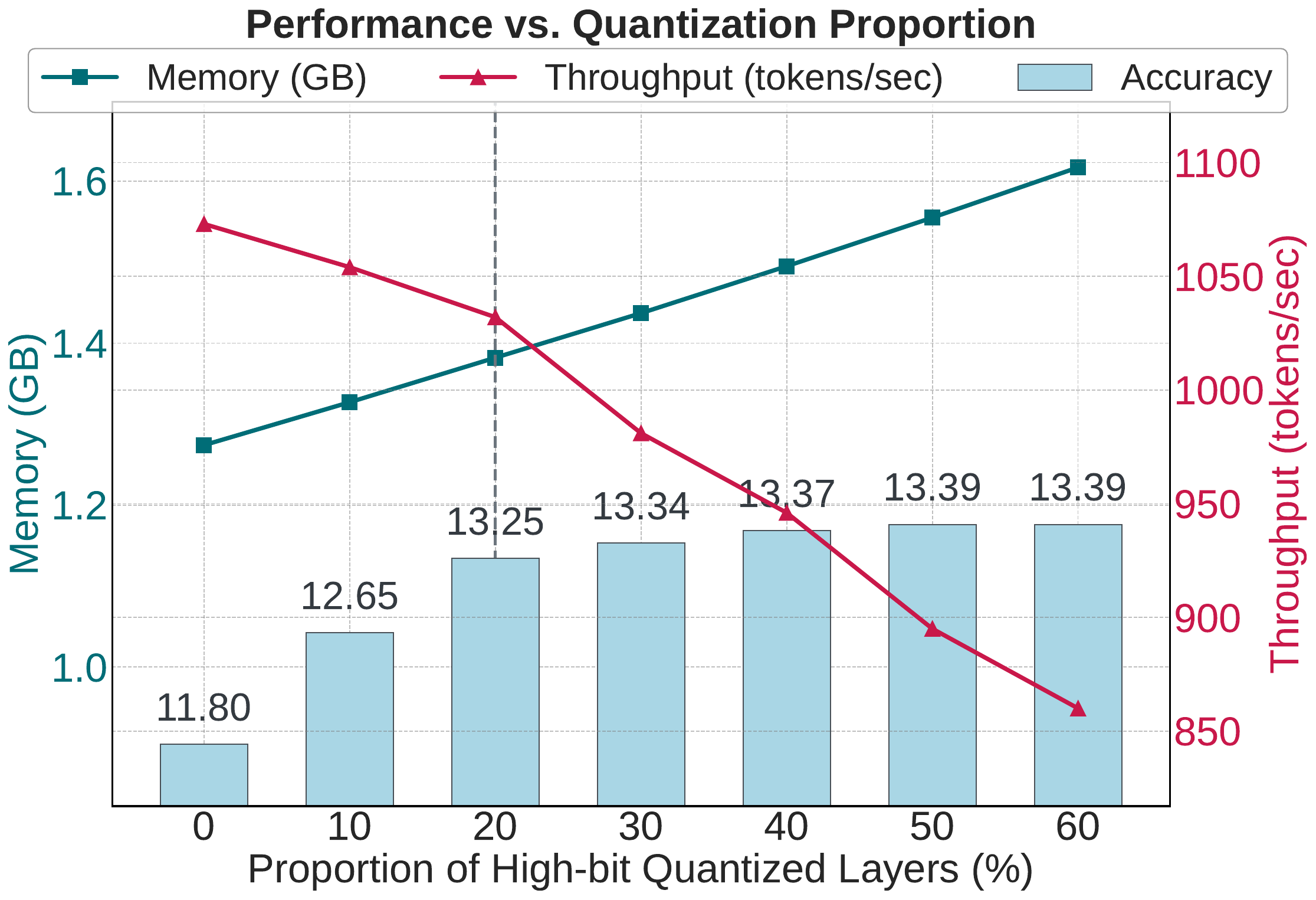}
	\caption{Performance variation of Llama 2-7B with different quantization configurations ("10\%" indicates the top 10\% important layers are quantized to 4 and 3 bits, and the remaining layers are quantized to 2 bits. The dataset is GSM8K).}
	\label{fig05}
\end{figure}

\begin{figure}[htbp]
	\centering
	\includegraphics[width=1.0\columnwidth]{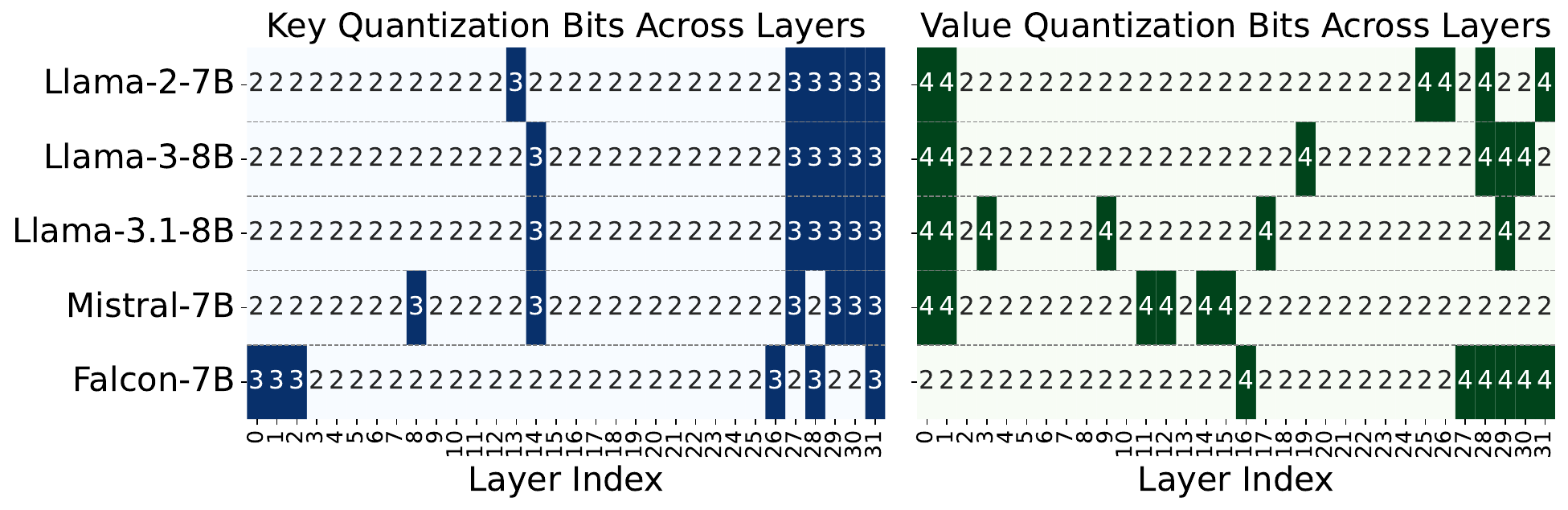}
	\caption{Detailed quantization configurations of KVmix-k2.19v2.38 for different models.}
	\label{fig06}
\end{figure}

We used 3-bit and 2-bit mixed quantization for Key, and 4-bit and 2-bit mixed quantization for Value. When the KV is quantized to 3 bits or 4 bits, the RPC proportion is set to 20\%; for 2-bit quantization, the RPC proportion is set to 10\%. Appendix C shows the impact of different RPC proportions on model performance. When the RPC proportion exceeds 20\%, its contribution to accuracy improvement is marginal; thus, we selected 20\% as the high-bit configuration for KVmix. The group size for quantization is 32. We selected 30 prompts from the LongBench for KV importance analysis. By the KVmix profiler, we can obtain the KV bit configurations and  RPC proportions for each layer. When utilizing the KVmix profiler, randomly selecting 20 to 30 prompts is sufficient to yield reliable importance analysis results (Appendix D provides analysis results derived from different prompts by the KVmix profiler); additional prompts do not significantly alter the outcomes. For the models and experimental environment used in this work, this process requires only 10 to 15 minutes, highlighting the efficiency of the KVmix profiler. Users can flexibly customize the quantization configuration by adjusting the proportion of layers with different bit-widths in the KVmix profiler to meet varying accuracy or memory requirements. Fig. \ref{fig05} illustrates the trends in accuracy, KV memory usage, and throughput as we varied the proportion of model layers quantized to 3 and 4 bits. When the proportion of model layers quantized to 3 and 4 bits is set to 20\%, the optimal tradeoff among these three factors is achieved. Under this configuration, the average quantization bit-width for Key is 2.19 (exact value: 2.1875), and for Value is 2.38 (exact value: 2.375). The detailed configuration obtained using the KVmix profiler is shown in Fig. \ref{fig06}. Unless otherwise specified, the configuration is applied to the k-2.19v2.38 quantization in subsequent experiments. 

%%Appendix C shows the impact of different RPC proportions on model performance.
%%(Appendix D provides analysis results derived from different prompts by the KVmix profiler)

\subsection{Performance Evaluation} \label{Model Performance Evaluation}

\subsubsection{Long Context} \label{Long Context Evaluation}

We evaluated the performance of various quantization configurations across 8 distinct datasets from the LongBench benchmark, using the FP16 model as the baseline. The detailed experimental results are presented in the Table \ref{table01}. The findings indicate that KVmix-k2.19v2.38 achieves an average accuracy loss (average values on 4 different models) of 1.67\% compared to the FP16 baseline. In contrast, the average accuracy loss of KVmix-2bit reached 4.53\%  relative to the baseline. In addition, random selection of high-bit quantization layers (random-k2.19v2.38) leads to an average accuracy loss of 4.06\%. These accuracy losses are significantly higher than KVmix-k2.19v2.38, demonstrating the advantage of KV importance-aware mixed quantization. The average accuracy of KVmix-k2.19v2.38w/oRPC decreases by 3.28\% compared with KVmix-k2.19v2.38, proving the effectiveness of using RPC for improving model accuracy. Appendix E presents the evaluation results on additional datasets and increasing the high-bit quantization layer to 30\% (KVmix-k2.28v2.56) on LongBench. KVmix-k2.28v2.56 nearly matches full 4-bit quantization performance but uses more memory than KVmix-k2.19v2.38.
%Appendix E presents the evaluation results on additional datasets and increasing the high-bit quantization layer to 30\% (KVmix-k2.28v2.56) on LongBench. KVmix-k2.28v2.56 nearly matches full 4-bit quantization performance but uses more memory than KVmix-k2.19v2.38.
\begin{table*}[!t]
\centering
\small 
% \scalebox{0.89}{
\setlength{\tabcolsep}{1.0mm}{
\begin{tabular}{l l c c c c c c c c c}
\toprule
\multirow{2}{*}{\textbf{Models}} & \multirow{2}{*}{\textbf{Methods}} & \multicolumn{8}{c}{\textbf{Datasets}} & \multirow{2}{*}{\textbf{Average}} \\
\cmidrule{3-10}
& & \textbf{\rotatebox{40}{TriviaQA}} & \textbf{\rotatebox{40}{Qasper}} & \textbf{\rotatebox{40}{MF-en}} & \textbf{\rotatebox{40}{QMSum}} & \textbf{\rotatebox{40}{2WikiMQA}} & \textbf{\rotatebox{40}{Rbench-P}} & \textbf{\rotatebox{40}{TREC}} & \textbf{\rotatebox{40}{PsgRetr-en}} & \\
\midrule
\multirow{4}{*}{Llama-2-7B} 
    & FP16             & 78.89 & 9.55  & 22.86 & 21.19 & 9.94  & 55.64 & 66.00   & 6.64  & 33.839 \\
    & KVmix-2bit       & 77.57 & 9.58  & 22.47 & 20.45 & 9.15  & 56.34 & 66.00   & 5.29  & 33.356 \\
    & random-k2.19v2.38& 78.30  & 9.39  & 22.54 & 20.41 & 9.46  & 56.36 & 66.00   & 5.49  & 33.494 \\
    & KVmix-k2.19v2.38w/oRPC& 77.95  & 9.19  & 21.03 & 19.98 & 9.05  & 56.13 & 65.50   & 5.61  & 33.055 \\
    %%\rowcolor{gray!20}
    &\cellcolor{gray!20}\textbf{KVmix-k2.19v2.38} & \cellcolor{gray!20}\textbf{78.78} & \cellcolor{gray!20}\textbf{9.59}  & \cellcolor{gray!20}\textbf{22.82} & \cellcolor{gray!20}\textbf{20.49} & \cellcolor{gray!20}\textbf{9.77}  & \cellcolor{gray!20}\textbf{56.54} & \cellcolor{gray!20}\textbf{66.00} & \cellcolor{gray!20}\textbf{5.72}  & \cellcolor{gray!20}\textbf{33.714} \\
\midrule
\multirow{4}{*}{Llama-3-8B} 
    & FP16 & 78.35 & 40.75 & 46.80  & 21.69 & 32.39 & 49.77 & 70.50 & 37.00    & 47.156 \\
    & KVmix-2bit       & 76.13 & 39.18 & 45.70  & 21.20  & 32.19 & 44.56 & 71.00   & 36.30  & 45.783  \\
    & random-k2.19v2.38& 78.01 & 39.17 & 45.90  & 21.22 & 32.02 & 45.36 & 71.00   & 36.50  & 46.148  \\
    & KVmix-k2.19v2.38w/oRPC& 77.12 & 39.04 & 45.18  & 21.03 & 32.05 & 45.20 & 71.00   & 36.00  & 45.828  \\
    %%\rowcolor{gray!20}
    & \cellcolor{gray!20}\textbf{KVmix-k2.19v2.38} & \cellcolor{gray!20}\textbf{78.13} & \cellcolor{gray!20}\textbf{39.15} & \cellcolor{gray!20}\textbf{46.31} & \cellcolor{gray!20}\textbf{21.26} & \cellcolor{gray!20}\textbf{32.20}  & \cellcolor{gray!20}\textbf{47.56} & \cellcolor{gray!20}\textbf{71.00}   & \cellcolor{gray!20}\textbf{36.50}  & \cellcolor{gray!20}\textbf{46.514} \\
\midrule
\multirow{4}{*}{Llama-3.1-8B} 
    & FP16 & 83.67 & 11.53 & 31.13 & 22.88 & 13.92 & 61.84 & 67.50 & 19.50  & 38.996 \\
    & KVmix-2bit & 83.10 & 10.90  & 30.76 & 22.11 & 13.08 & 58.92 & 67.00   & 19.00    & 38.109 \\
    & random-k2.19v2.38& 83.25 & 10.90  & 31.05 & 22.34  & 13.05 & 59.26 & 67.00 & 19.00  & 38.231 \\
    & KVmix-k2.19v2.38w/oRPC& 82.18 & 11.05  & 30.86 & 22.20  & 13.27 & 58.51 & 67.00 & 19.00  & 38.009 \\
    %%\rowcolor{gray!20}
    & \cellcolor{gray!20}\textbf{KVmix-k2.19v2.38} & \cellcolor{gray!20}\textbf{83.28} & \cellcolor{gray!20}\textbf{11.40} & \cellcolor{gray!20}\textbf{31.49}  & \cellcolor{gray!20}\textbf{22.90} & \cellcolor{gray!20}\textbf{12.92} & \cellcolor{gray!20}\textbf{59.96} & \cellcolor{gray!20}\textbf{67.50} & \cellcolor{gray!20}\textbf{19.50}  & \cellcolor{gray!20}\textbf{38.619} \\
\midrule
\multirow{4}{*}{Mistral-7Bv0.3} 
    & FP16 & 84.29 & 36.19 & 54.70 & 21.79 & 35.08 & 53.06 & 73.50 & 32.50 & 48.889  \\
    & KVmix-2bit & 84.08 & 34.29 & 53.87 & 21.37 & 33.39 & 50.99 & 73.50 & 32.00    & 47.936 \\
    & random-k2.19v2.38& 84.01 & 34.35 & 53.61 & 21.45 & 33.40 & 50.59 & 73.50 & 32.50  & 47.926 \\ 
    & KVmix-k2.19v2.38w/oRPC& 83.07 & 34.18 & 52.65 & 21.10 & 32.32 & 51.30 & 73.50 & 32.50  & 47.578 \\
    %%\rowcolor{gray!20}
    & \cellcolor{gray!20}\textbf{KVmix-k2.19v2.38} & \cellcolor{gray!20}\textbf{84.03} & \cellcolor{gray!20}\textbf{35.67} & \cellcolor{gray!20}\textbf{53.68} & \cellcolor{gray!20}\textbf{21.84} & \cellcolor{gray!20}\textbf{33.81} & \cellcolor{gray!20}\textbf{51.98} & \cellcolor{gray!20}\textbf{73.50} & \cellcolor{gray!20}\textbf{32.75} & \cellcolor{gray!20}\textbf{48.408}  \\
\midrule
\multirow{4}{*}{Falcon-7B} 
    & FP16 & 6.94 & 3.87 & 7.47 & 3.96 & 4.87 & 12.92 & 14.00 & 3.95 & 7.248  \\
    & KVmix-2bit & 5.96 & 3.15 & 6.19 & 3.28 & 4.22 & 11.40 & 13.50 & 3.21    & 6.364 \\
    & random-k2.19v2.38& 6.11 & 3.10 & 6.36 & 3.26 & 4.24 & 11.45 & 13.50 & 3.26  & 6.410 \\ 
    & KVmix-k2.19v2.38w/oRPC& 6.05 & 3.01 & 6.54 & 3.22 & 4.13 & 11.41 & 13.00 & 3.22  & 6.323 \\
    %%\rowcolor{gray!20}
    & \cellcolor{gray!20}\textbf{KVmix-k2.19v2.38} & \cellcolor{gray!20}\textbf{6.64} & \cellcolor{gray!20}\textbf{3.28} & \cellcolor{gray!20}\textbf{7.06} & \cellcolor{gray!20}\textbf{3.52} & \cellcolor{gray!20}\textbf{4.60} & \cellcolor{gray!20}\textbf{12.71} & \cellcolor{gray!20}\textbf{14.00} & \cellcolor{gray!20}\textbf{3.62} & \cellcolor{gray!20}\textbf{6.929}  \\
\bottomrule
\end{tabular}}
    %}
    \caption{Model accuracy of 4 LLMs on LongBench with different quantization configurations. KVmix-k2.19v2.38 uses the configurations of Fig. \ref{fig06}. KVmix-2bit uses the asymmetric 2-bit (Key per-channel and Value per-token) quantization for all model layers (RPC ratio is set to 10\%). random-k2.19v2.38 randomly selects 20\% of the model layers to perform asymmetric 3-bit and 4-bit quantization for Key and Value (RPC ratio is set to 20\%), and the remaining layers are 2-bit quantization (RPC ratio is set to 10\%). KVmix-k2.19v2.38w/oRPC is KVmix-k2.19v2.38 without RPC (RPC ratio is set to 0\%).}
    \label{table01}
\end{table*}

We compared KVmix against prior SOTA methods for KV Cache, specifically Key per-channel and Value per-token methods, i.e., KIVI \cite{liu2024kivi} and KVQuant \cite{hooper2024kvquant}, since they are known to minimize KV quantization errors. Additionally, we evaluated KVmix against the latest SOTA method, QJL \cite{zandieh2025qjl}. Table \ref{table03} presents the accuracy results. The results show that KVmix-k2.19v2.38 surpasses the performance of KIVI-2bit-r64 and QJL-3bit, reducing the average accuracy loss by 1.50\% and 0.68\%, respectively. While KVQuant-3bit-1\% achieves accuracy comparable to KVmix-k2.19v2.38, its memory compression ratio and inference efficiency fall short of those delivered by KVmix-k2.19v2.38 (Fig. \ref{fig07}, Fig. \ref{fig08}). By increasing the quantization bit-width, KVmix-k2.28v2.56 demonstrates a more significant advantage over KVQuant-3bit-1\% in accuracy, while maintaining a comparable memory compression ratio (4.8$\times$) and superior inference acceleration (5.23$\times$). This flexibility in balancing accuracy and quantization bit-width represents a critical strength of KVmix.

\begin{table*}[htbp]
    \centering
    \small 
    %%\scalebox{0.88}{
   \setlength{\tabcolsep}{1mm}{
    \begin{tabular}{lccccccccc}
        \toprule
        \textbf{Methods} & \textbf{TriviaQA} & \textbf{Qasper} & \textbf{MF-en} & \textbf{QMSum} & \textbf{2WikiMQA} & \textbf{Repobench-P} & \textbf{TREC} & \textbf{PsgRetr-en} & \textbf{Average} \\
        \midrule
        FP16 & 78.89 & 9.55 & 22.86 & 21.19 & 9.94 & 55.64 & 66.00 & 6.64 & 33.839 \\
        KIVI-2bit-r64 & 77.08 & 9.16 & 22.55 & 20.12 & 9.05 & 56.15 & 66.00 & 5.62 & 33.216 \\
        QJL-3bit & 78.25 & 9.10 & 22.60 & 20.45 & 9.68 & 56.09 & 66.00 & 5.72 & 33.486 \\
        KVQuant-3bit-1\% & 78.79 & 10.51 & 22.61 & 20.58 & 9.75 & 55.62 & 66.00 & 5.76 & 33.703 \\
        \rowcolor{gray!20}
        KVmix-k2.19v2.38 & 78.78 & 9.59 & 22.82 & 20.49 & 9.77 & 56.54 & 66.00 & 5.72 & \textbf{33.714} \\
        \rowcolor{gray!20}
        KVmix-k2.28v2.56 & 78.05 & 10.21 & 23.21 & 20.63 & 9.72 & 56.61 & 66.00 & 6.08 & \textbf{33.814} \\
        \bottomrule
    \end{tabular}}%%}
        \caption{Accuracy comparison of different quantization methods on LongBench using the Llama 2-7B-hf model. KIVI-2bit-r64 uses 2-bit quantization with a full-precision residual of 64. KVQuant-3bit-1\% uses 3-bit quantization and 1\% outlier handling. QJL-3bit uses 3-bit quantization. KVmix-k2.28v2.56 increases the proportion of high-bit quantization layers in KVmix-k2.19v2.38 to 30\%, its detailed configurations are shown in Appendix E.}
        \label{table03}
\end{table*}
%%, its detailed configurations are shown in Appendix E

\subsubsection{GSM8K and Wikitext-2} \label{GSM8K and Wikitext-2 Evaluation}

We evaluated the capabilities of the quantized model in language modeling and mathematical reasoning using the FP16 model as the baseline. Atom \cite{zhao2024atom} exhibits very poor performance in long contexts, and thus, we only compare it in this section. The evaluation was conducted using the lm\_eval \cite{eval-harness} framework, where the quantized model replaced the Hugging Face model. Specifically, we measured the accuracy on the GSM8K dataset and the perplexity on the Wikitext-2 dataset. The experimental results are detailed in Table \ref{table02}. The results show that 2bit (k-T, v-T) suffers a catastrophic performance loss on GSM8K and Wikitext-2, and the model almost loses its reasoning ability. For 4bit (k-T, v-T), the performance loss of the model on GSM8K and Wikitext-2 also reached 9.17\% and 5.28\%, respectively. In contrast, on the Wikitext-2, the perplexity score of KVmix-k2.19v2.38 is almost comparable to the baseline, while on the more challenging GSM8K mathematical reasoning task, KVmix-k2.19v2.38 has an accuracy loss of 2.00\%, which significantly outperforms the 2bit (k-T, v-T) and the 4bit (k-T, v-T). Moreover, on GSM8K, KVmix-k2.19v2.38 shows a significant accuracy improvement compared to KVmix-2bit and random-k2.19v2.38, which do not leverage the KV importance analysis for more accurate quantization. Compared to Atom-4bit and other SOTA methods, KVmix-k2.19v2.38 also has an accuracy advantage. Notably, the Atom-4bit performs 4-bit quantization on both the model weights and activations, which results in greater accuracy loss. These results demonstrate the superior performance of KVmix-k2.19v2.38 in complex task reasoning.

\begin{table}[htbp]
%%\begin{table}[htbp]
    \centering
    \small 
     \setlength{\tabcolsep}{1mm}{
    \begin{tabular}{lcc}
        \toprule
        \textbf{Methods} & \textbf{GSM8K (acc$\uparrow$)} & \textbf{Wikitext-2 (ppl$\downarrow$)} \\
        \midrule
         FP16 & 13.52 & 8.71 \\
         2bit (k-T, v-T) & 0.83 & 11089 \\
         4bit (k-T, v-T) & 12.28 & 9.17 \\
         KVmix-2bit & 11.80 & 8.73 \\
         random-k2.19v2.38 & 11.97 & 8.73 \\
         Atom-4bit & 12.30 & 9.32 \\
         KIVI-2bit-r64 & 12.75 & 8.80 \\
         QJL-3bit & 13.11 & 8.75 \\
         KVQuant-3bit-1\% & 13.23 & 8.71 \\
        %%\rowcolor{gray!20}
         \cellcolor{gray!20}\textbf{KVmix-k2.19v2.38} & \cellcolor{gray!20}\textbf{13.25} & \cellcolor{gray!20}\textbf{8.71} \\
        \bottomrule
    \end{tabular}}
%%\end{table}
    \caption{Model accuracy (acc) on GSM8K and perplexity (ppl) on Wikitext-2 using Llama 2-7B-hf. 2bit (k-T, v-T) uses the symmetric 2-bit (Key per-token and Value per-token) quantization for all model layers, and 4bit (k-T, v-T) uses the symmetric 4-bit quantization; their RPC ratio is set to 0.}
    \label{table02}
\end{table}
 
\subsection{Inference Efficiency and Memory Usage Evaluation} \label{Model Efficiency and Memory Usage Evaluation}

\begin{figure}[htbp]
	\centering
	\includegraphics[width=1.0\columnwidth]{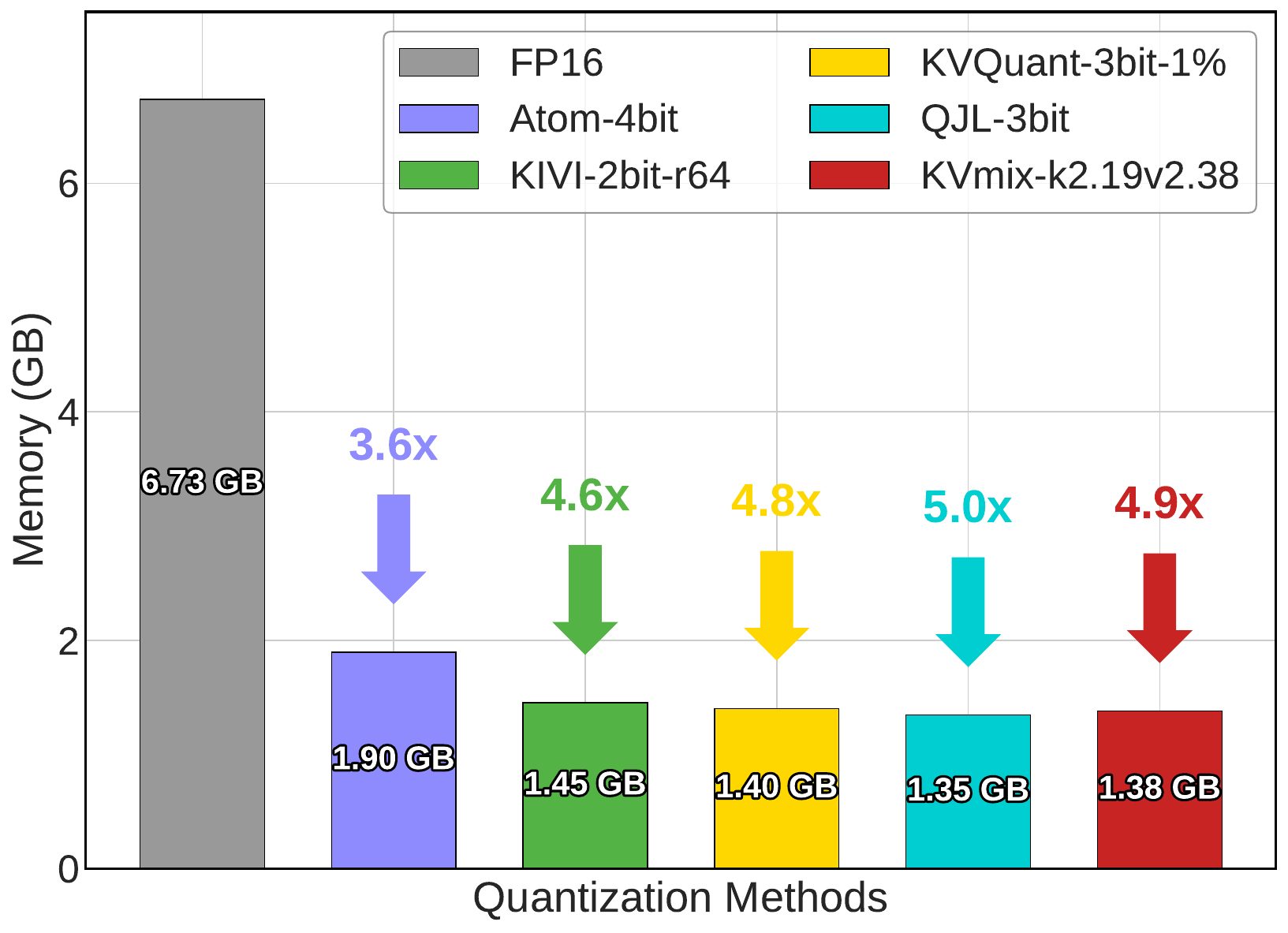}
	\caption{Dynamic peak memory usage of different methods during inference on the Llama 2-7B-hf model.}
	\label{fig07}
\end{figure}

\begin{figure}[htbp]
	\centering
	\includegraphics[width=1.0\columnwidth]{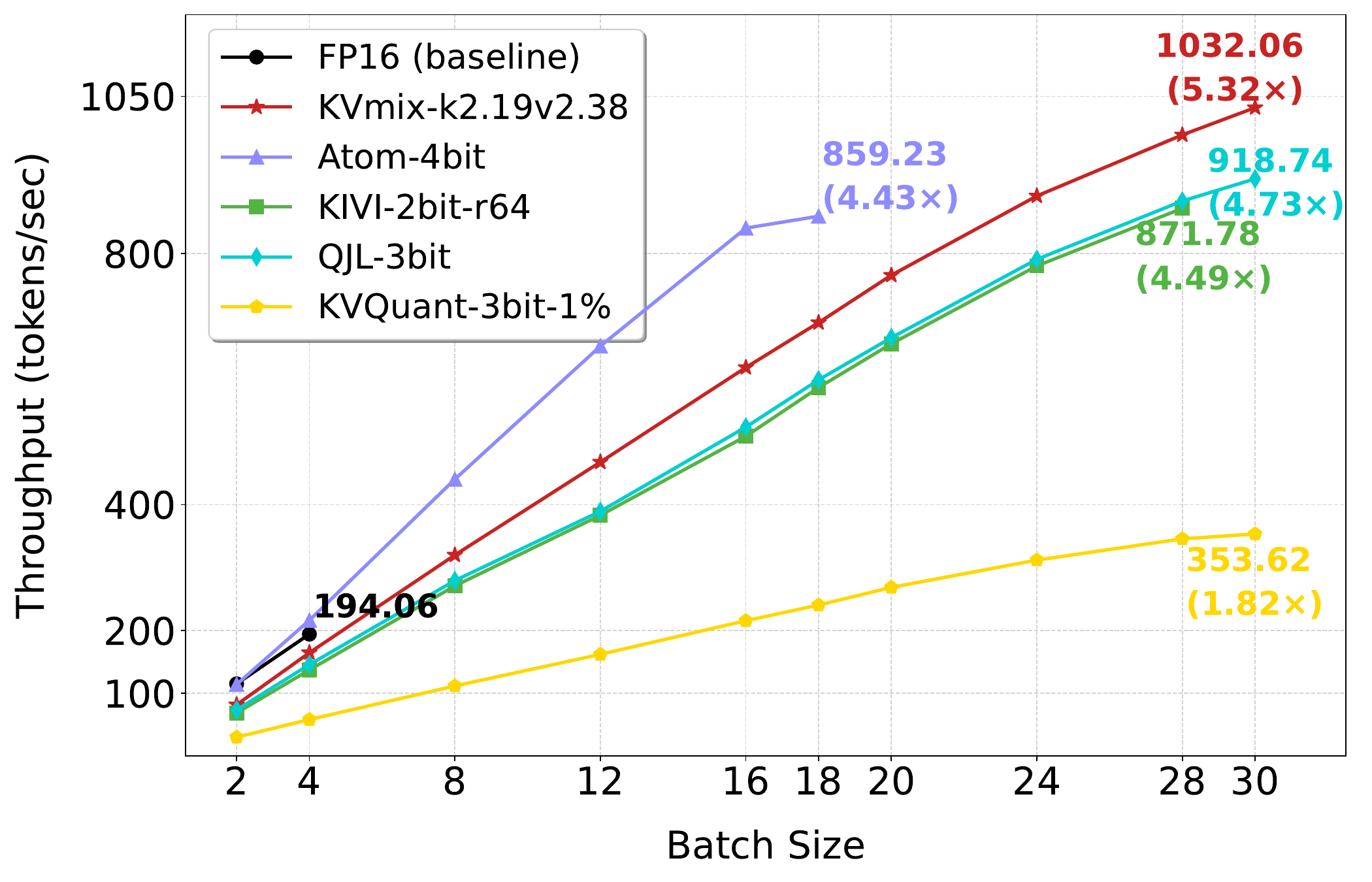}
	\caption{Inference throughput of different quantization methods with different batch sizes on the Llama 2-7B-hf model.}
	\label{fig08}
\end{figure}

We evaluated the inference throughput and memory usage of KVmix during inference. To ensure fairness, we applied identical input data across all evaluated methods. The number of input tokens is 688, the maximum number of new tokens is set to 1024, and the model is Llama 2-7B-hf. We compared KVmix against the KIVI-2bit-r64, KVQuant-3bit-1\%, QJL-3bit, and Atom-4bit. Memory usage results are illustrated in Fig. \ref{fig07}, with a batch size fixed at 4. The reported memory usage represents the peak memory usage during inference minus the memory occupied by the model before inference. To fully utilize the GPU memory, we incrementally increased the batch size to explore KVmix’s maximum throughput. The throughput results are shown in Fig. \ref{fig08}. The baseline (FP16), Aotm-4bit, and KIVI-2bit-r64 reach out of memory at batch sizes of 4, 18, and 28, respectively, while the KVmix-k2.19v2.38 can reach a maximum batch size of 30 with an inference throughput of 1032 tokens per second.
%%We compared KVmix against the methods optimized for memory and throughput: KIVI \cite{liu2024kivi} with 2-bit quantization (full precision residual of 64) and Atom \cite{zhao2024atom} with 4-bit quantization.

The results reveal that KVmix-k2.19v2.38 achieves a 4.9$\times$ reduction in memory usage and up to a 5.3$\times$ increase in throughput compared to the baseline. This efficient memory compression stems from KVmix’s extremely low bit quantization and dynamic RPC strategy, which progressively reduces the full-precision KV as inference progresses. In contrast, KIVI employs a fixed full-precision residual strategy, unable to dynamically reduce the number of full-precision KVs. Thus, KVmix saves more memory than KIVI-2bit despite using Key-2.19 and Value-2.38 bit quantization. Meanwhile, Atom quantizes both model weights and activations while utilizing tensor cores for optimized kernel, achieving a higher throughput at the same batch size but incurring greater model accuracy degradation (Table \ref{table02}). While KVQuant achieves significant memory compression, its inference efficiency is hampered by substantial preprocessing requirements. QJL implements ``zero-overhead'' quantization by eliminating the need to store extra constants like zero-points and scaling factors. This allows it to achieve a slightly better memory compression compared to KVmix, but its inference efficiency and accuracy are lower than those of KVmix.

\section{Conclusion}

This paper proposes KVmix, a novel mixed quantization method tackling the KV Cache memory bottleneck in LLM inference. KVmix creatively integrates layer importance analysis based on KV weight gradients into KV quantization and integrates dynamic long-context optimization to cut memory usage while maintaining generation quality. It achieves significant memory and efficiency gains with minimal loss in accuracy, offering flexibility to adapt quantization strategies to diverse scenarios. Future work will explore integrating lightweight mechanisms for real-time KV bit adjustments into KVmix to enhance adaptability.

\section*{Acknowledgments}
This research was funded by National Key R\&D Program of China (2022YFB4501604).

\bibliography{li-Reference.bib}

\section{Appendix A}
\subsection{KV Weight Analysis}

\begin{figure*}[htbp]
	\centering
        \includegraphics[width=1.0\linewidth]{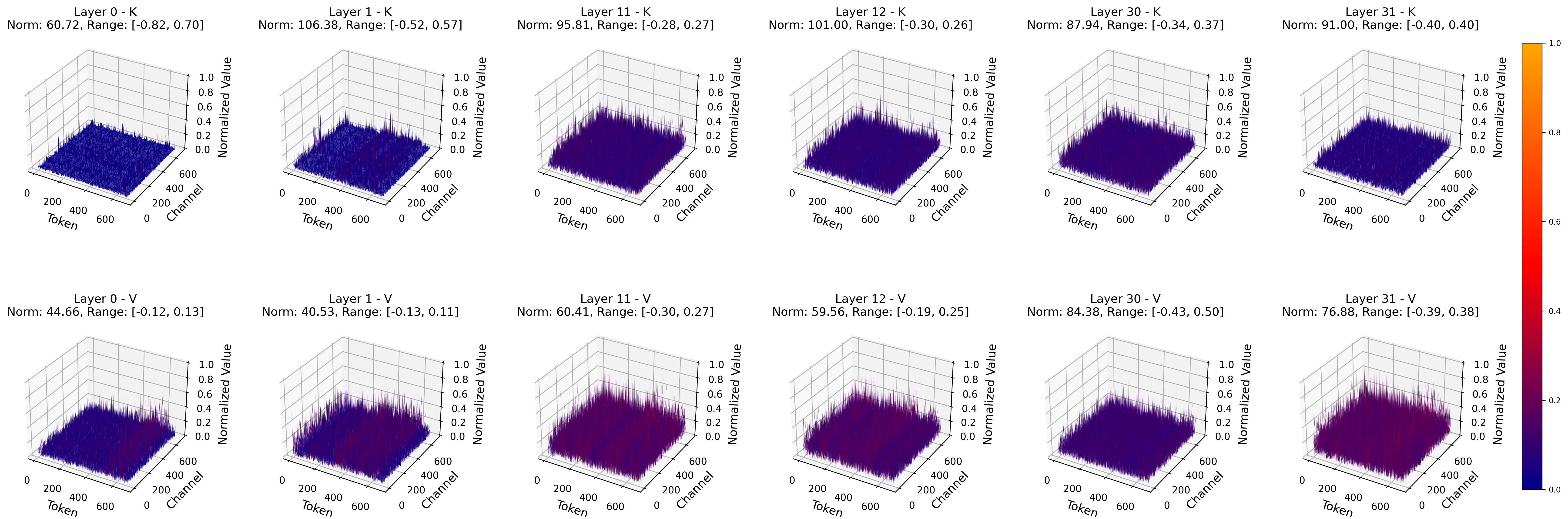}
        \includegraphics[width=1.0\linewidth]{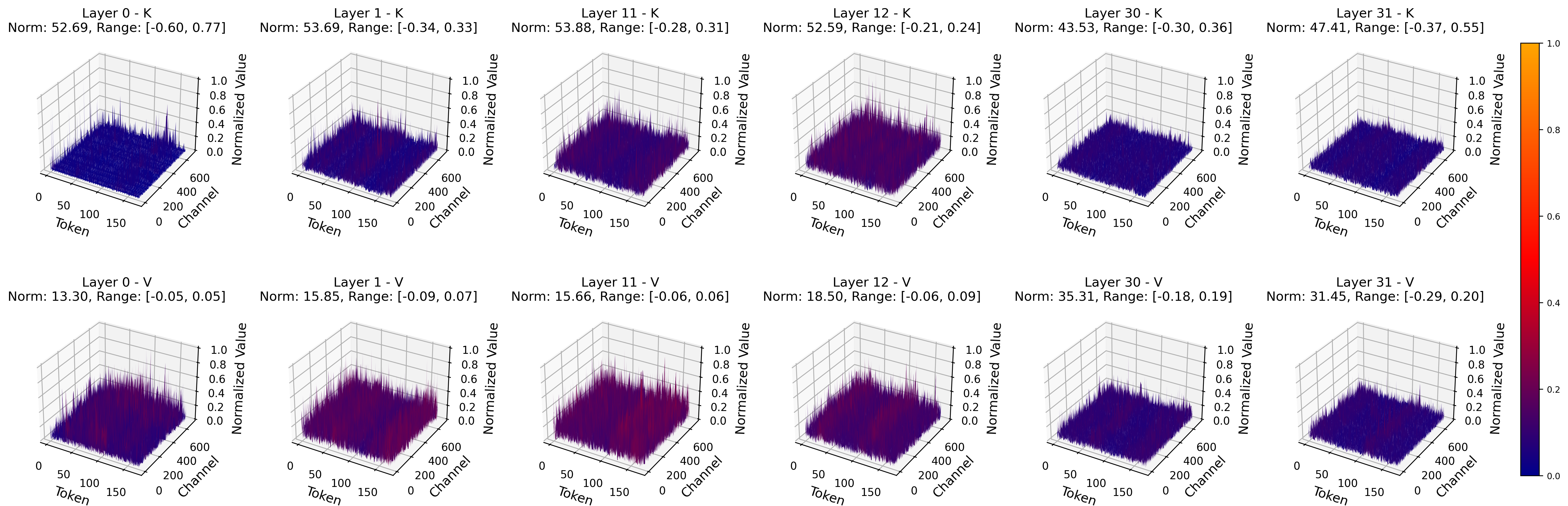}
        \includegraphics[width=1.0\linewidth]{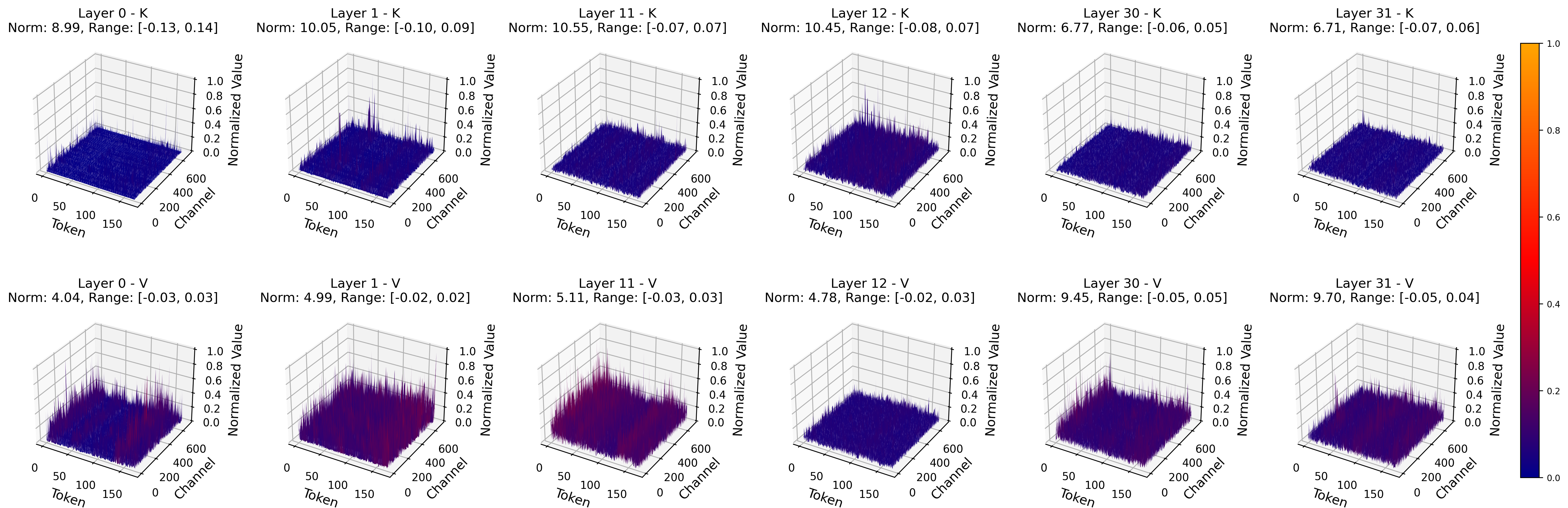}
	\caption{Projection matrix weights of K and V across different model layers for Llama 2-7B-hf (top), Llama 3-8B-Instruct (middle) and Mistral-7B-Instruct-v0.3 (bottom). The heatmap visualizes the projection weights by sampling every sixth element in the matrices. Additionally, the weight values along the axes have been normalized. ``Norm" represents the L2 norm of the weight matrix for each layer, while ``Range" indicates the range of values within each layer's weight matrix.}
	\label{fig09}
\end{figure*}

Fig. \ref{fig09} provides a comprehensive heatmap analysis of the Key ($W_k$) and Value ($W_v$) projection weight matrices across different layers for three models: Llama2-7B-hf, Llama 3-8B-Instruct, and Mistral-7B-Instruct-v0.3. Through a multi-model comparison, Fig. 1 reveals the detailed characteristics of the weight distributions for $W_k$ and $W_v$. The heatmaps highlight the following observations: there are significant differences in the magnitude and distribution of $W_k$ and $W_v$ across layers in different models, and within the same layer, the weight patterns vary distinctly across models. These findings further validate two key conclusions: (1) the variability of KV weights across layers suggests the need for layer-specific quantization strategies rather than a uniform low-bit quantization strategy; (2) the difference in Key and Value weights distribution within the same layer indicates the need to use different quantization bits for Key and Value. This analysis provides a foundation for the layer importance-aware quantization method, demonstrating the necessity of customizing quantization bit-widths for different layers and models.

%%\ding{172} \textit{Significant variations in KV weight values across different layers.} \ding{173} \textit{Distinct distribution patterns of KV weights within the same layer.}

\section{Appendix B}
\subsection{KVmix Profiler Algorithm}
In Algorithm \ref{algorithm01}, we present the algorithm workflow of the KVmix profiler, which encompasses three core steps: data processing, KV importance analysis, and quantization parameters configuration. This process is performed offline. By computing the gradient norms of the model's loss function with respect to $W_k$ and $W_v$, the importance of KV at each layer is assessed. Once the layer importance analysis and quantization parameters configuration are completed, the quantization results can be directly reused for subsequent inference tasks without the need for recalculation, thereby ensuring that inference efficiency remains unaffected. Although the analysis involves backpropagation, which demands significant memory, this pressure can be effectively mitigated by limiting the length of input prompts. Selecting a sufficient number of prompts (e.g., 20 to 30) ensures that this optimization does not significantly impact the accuracy of the layer importance analysis.

\begin{breakablealgorithm}
\caption{Model Profiling and Quantization}
\label{algorithm01}
\begin{algorithmic}[1]
\Require {Model $model$, inputs $input\_ids$, $attention\_mask$, $labels$, layers $num\_layers$, datasets $datasets$}
\Ensure {Quantization bits $quant\_bits$}
\For{\textit{dataset} in \textit{datasets}}
    \State \textit{prompts} $\gets$ randomSample(dataset, \textit{n}); \quad $\triangleright$ Sample \textit{n} prompts
    \State \textit{inputs} $\gets$ tokenize(\textit{prompts}); \quad $\triangleright$ Prepare inputs
    \State \textit{model} $\gets$ loadFullModel(); \quad $\triangleright$ Load full-precision model
    \State \textit{kScores, vScores} $\gets$ \textit{calcImportance}(\textit{model, inputs, num\_layers}); \quad $\triangleright$ Compute importance
    \State \textit{kLayers, vLayers} $\gets$ \textit{classify}(\textit{kScores, vScores, num\_layers}); \quad $\triangleright$ Classify layers
    \State \textit{quantBits} $\gets$ \textit{setBits}(\textit{kLayers, vLayers}); \quad $\triangleright$ Set quantization bits
    \State del \textit{model}; \quad $\triangleright$ Free memory
    \State \textit{model} $\gets$ loadQuantModel(\textit{quantBits}); \quad $\triangleright$ Load quantized model
    \State \textit{preds} $\gets$ infer(\textit{model, dataset}); \quad $\triangleright$ Inference
    \State save(\textit{preds}); \quad $\triangleright$ Save results
    \State del \textit{model}; \quad $\triangleright$ Free memory
\EndFor

\State \textbf{Function} \textit{calcImportance}(\textit{model, inputs, num\_layers}): \quad $\triangleright$ Compute importance scores
    \State \textit{kScoresAll, vScoresAll} $\gets$ [], [];
    \For{\textit{input} in \textit{inputs}}
        \State \textit{kScores, vScores} $\gets$ [], [];
        \For{\textit{layer} in range(\textit{num\_layers})}
            \State \textit{loss} $\gets$ \textit{model.forward}(\textit{input}); \quad $\triangleright$ Compute loss
            \State \textit{kGrad} $\gets$ grad(\textit{loss, kParams(layer)}); \quad $\triangleright$ Gradient for Key
            \State \textit{vGrad} $\gets$ grad(\textit{loss, vParams(layer)}); \quad $\triangleright$ Gradient for Value
            \State \textit{kScores.append}(\textit{kGrad.norm()}); \quad $\triangleright$ Append Key score
            \State \textit{vScores.append}(\textit{vGrad.norm()}); \quad $\triangleright$ Append Value score
            \State clearMemory(); \quad $\triangleright$ Free memory
        \EndFor
        \State \textit{kScoresAll.append}(\textit{kScores});
        \State \textit{vScoresAll.append}(\textit{vScores});
    \EndFor
    \State \textit{kScoresMean} $\gets$ mean(\textit{kScoresAll}); \quad $\triangleright$ Average Key scores
    \State \textit{vScoresMean} $\gets$ mean(\textit{vScoresAll}); \quad $\triangleright$ Average Value scores
    \State \textbf{return} \textit{kScoresMean, vScoresMean};
\end{algorithmic}
\end{breakablealgorithm}

\section{Appendix C} \label{Appendix C}
\subsection{Analysis Results of KVmix Profiler Using Different Prompts}

\begin{figure*}[htbp]
	\centering
	\includegraphics[width=0.90\linewidth]{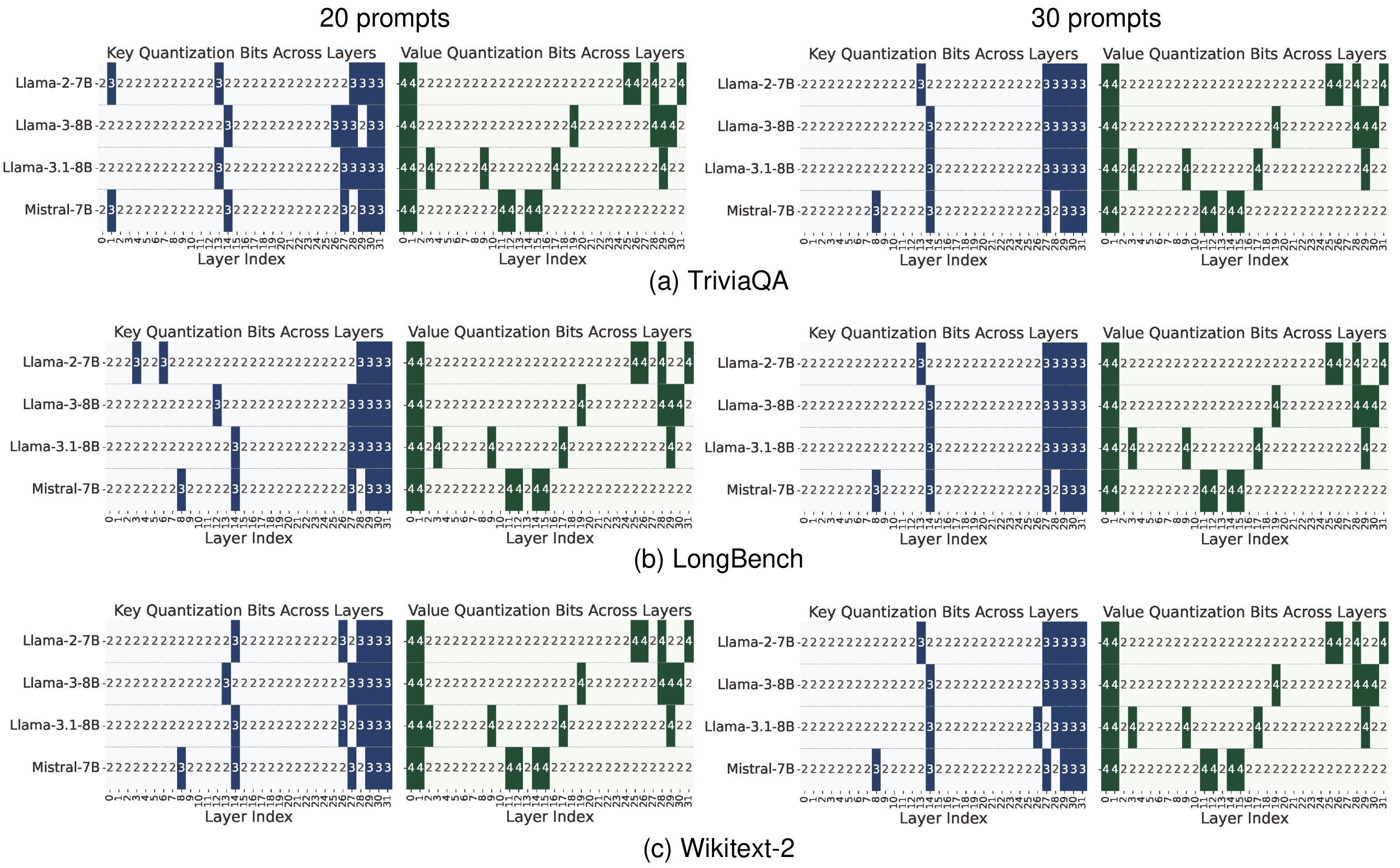}
	\caption{KV configuration results from the KVmix profiler using different datasets and prompts (20\% of the layers quantized to 3 bits or 4 bits, and the remaining layers quantized to 2 bits).}
	\label{fig12}
\end{figure*}

To verify the robustness of the KVmix profiler, we evaluated its layer importance analysis using various sets of prompts. These included the first 20 and 30 prompts from the LongBench TriviaQA dataset, 20 and 30 prompts randomly selected from LongBench, and 20 and 30 prompts randomly chosen from the Wikitext-2 dataset. The results, as shown in Fig. \ref{fig12}, demonstrate that the distribution of layer importance remains highly consistent regardless of the source or number of prompts used. This consistency is primarily due to our reliance on the model’s layer KV weights ($W_k$ and $W_v$) as the basis for analyzing KV importance; since the $W_k$ and $W_v$ remain unchanged during inference, a sufficient number of prompts is adequate to yield reliable importance analysis results. Furthermore, the analysis using 20 prompts yields results nearly identical to those obtained with 30 prompts, with the differences having a negligible impact on the model's final inference accuracy. This stability confirms the reliability of the KVmix profiler across different input scenarios.

\section{Appendix D}
\subsection{Impact of RPC Proportion on Model Accuracy}

\begin{figure*}[htbp]
	\centering
	\includegraphics[width=0.8\linewidth]{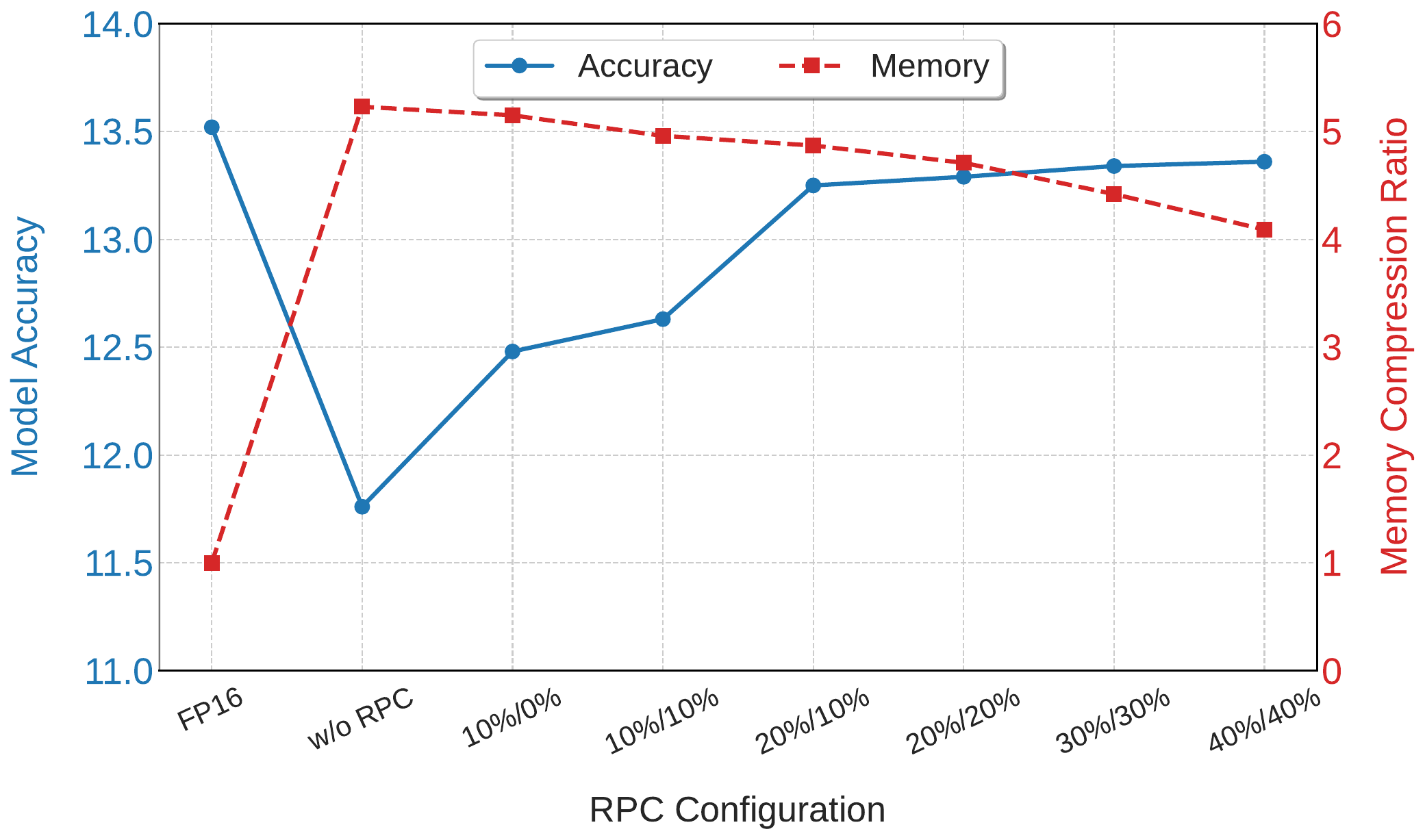}
	\caption{Accuracy and memory compression ratio variations of KVmix-k2.19v2.38 on Llama 2-7B with GSM8K dataset using different RPC ratios.}
	\label{fig13}
\end{figure*}

We systematically evaluated the impact of the Recent Pivotal Context (RPC) proportion on model accuracy by adjusting its proportion in the KVmix-k2.19v2.38 configuration, with results presented in Table \ref{table011}. The experiments demonstrate that when the RPC ratio is set to 0\% (without using RPC), the model's accuracy decreases by 2.67\% compared to the FP16 baseline. In contrast, when employing a 20\%/10\% RPC ratio, the accuracy only drops by 0.86\% relative to the baseline, highlighting the importance of RPC in preserving model accuracy. However, when the RPC ratio exceeds 20\%, further increases in the ratio result in only marginal improvements in model accuracy. Additionally, in short-context scenarios, where fewer new KV pairs are generated during the decoding phase, an excessively high RPC ratio results in a large proportion of full-precision KVs, thereby diminishing the memory compression benefits of quantization. As illustrated in Fig. \ref{fig13}, we evaluated the impact of varying RPC ratios on model accuracy and memory compression ratio using the KVmix-k2.19v2.38 configuration on the GSM8K dataset. The results indicate that when the RPC ratio for critical layers exceeds 20\%, the model accuracy remains nearly constant, while the memory compression ratio experiences a substantial decline. Therefore, it is recommended to limit the RPC ratio to within 20\% when configuring the KVmix profiler to achieve a balance between maintaining inference accuracy and avoiding excessive memory pressure.

\begin{table*}[htbp]
\centering
\small
\scalebox{1.0}{
\setlength{\tabcolsep}{1.0mm}{
\begin{tabular}{l c c c c c c c c c c c}
\toprule
 \multirow{2}{*}{\textbf{Methods}} & \multicolumn{10}{c}{\textbf{Datasets}} & \multirow{2}{*}{\textbf{Average}} \\
\cmidrule{2-11}
 & \textbf{\rotatebox{50}{TriviaQA}} & \textbf{\rotatebox{50}{Qasper}} & \textbf{\rotatebox{50}{MF-en}} & \textbf{\rotatebox{50}{QMSum}} & \textbf{\rotatebox{50}{2WikiMQA}} & \textbf{\rotatebox{50}{GovRep}} & \textbf{\rotatebox{50}{Rbench-P}} & \textbf{\rotatebox{50}{LCC}} & \textbf{\rotatebox{50}{TREC}} & \textbf{\rotatebox{50}{PsgRetr-en}} & \\
\midrule
% \multirow{6}{*}{Llama-2-7B-hf} 
     FP16  & 78.89 & 9.55  & 22.86 & 21.19 & 9.94  & 17.36 & 55.64 & 66.70 & 66.00 & 6.64  & 35.477 \\
     w/oRPC & 77.95 & 9.19  & 21.03 & 19.98 & 9.05  & 15.10 & 56.13 & 66.00 & 65.50 & 5.61  & 34.554 \\
     10\%/0\% & 78.20 & 9.29  & 21.55 & 19.90 & 9.33  & 15.10 & 56.44 & 66.30 & 66.00 & 5.60  & 34.771 \\
     10\%/10\% & 78.20 & 9.35  & 21.53 & 20.15 & 9.35  & 15.10 & 56.40 & 66.35 & 66.00 & 5.65  & 34.808 \\
     \cellcolor{gray!20}{20\%/10}\% & \cellcolor{gray!20}{78.78} & \cellcolor{gray!20}{9.59}  & \cellcolor{gray!20}{22.82} & \cellcolor{gray!20}{20.49} & \cellcolor{gray!20}{9.77} & \cellcolor{gray!20}{15.45} & \cellcolor{gray!20}{56.54} & \cellcolor{gray!20}{66.59} & \cellcolor{gray!20}{66.00} & \cellcolor{gray!20}{5.72}  & \cellcolor{gray!20}{35.175} \\
    %%\rowcolor{gray!20}
     20\%/20\% & 78.78 & 9.63 & 22.80 & 20.55 & 9.77  & 15.43 & 56.55  & 66.50 & 66.00  & 5.78  & 35.179 \\
    %%\rowcolor{gray!20}
     30\%/30\% & 78.02 & 10.23 & 23.21 & 20.63  & 9.75 & 15.60 & 56.61 & 66.70 & 66.00 & 5.98  & 35.273 \\
\bottomrule
\end{tabular}}
}
\caption{Model accuracy of Llama 2-7B on LongBench under different RPC ratios. ``w/o RPC": Indicates that the RPC ratio is set to 0 for all layers. ``10\%/0\%": Indicates that the RPC ratio is set to 10\% for Keys and Values quantized at 3-bit and 4-bit, and 0\% for Keys and Values quantized at 2-bit, with similar interpretations for other ratios.}
\label{table011}
\end{table*}

\section{Appendix E}
\subsection{Additional Performance Evaluation of KVmix on LongBench}
\begin{figure*}[htbp]
	\centering
	\includegraphics[width=1.0\linewidth]{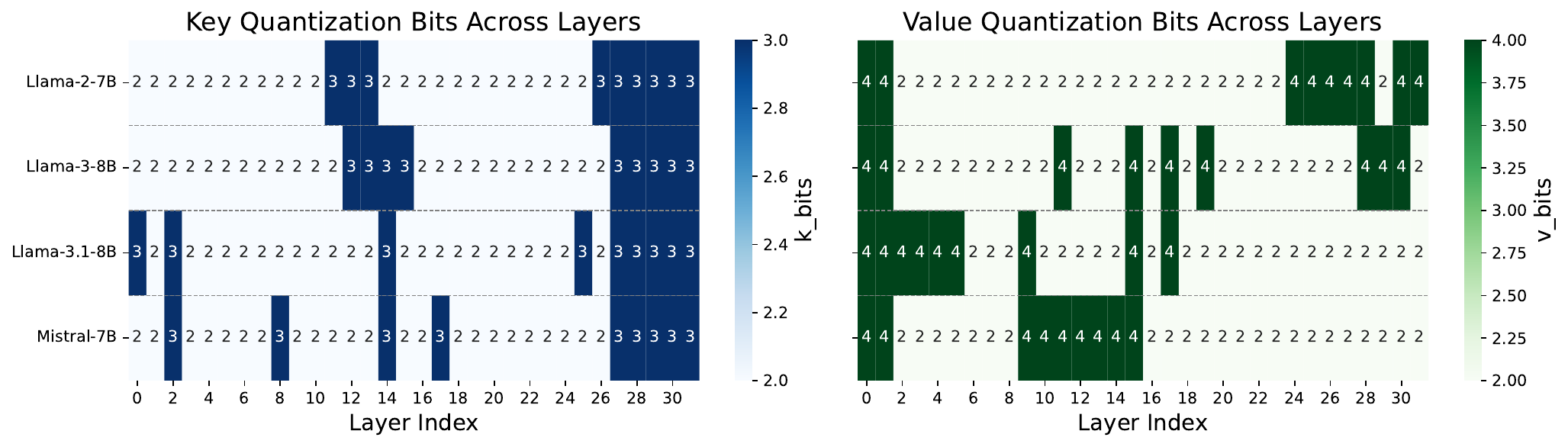}
	\caption{Detailed quantization configuration of KVmix-k2.28v2.56 in different models.}
	\label{fig14}
\end{figure*}

In this section, we present extended experimental results on LongBench to provide a more comprehensive performance comparison and to elucidate the impact of different KVmix configurations on model accuracy. Fig. \ref{fig14} illustrates the detailed configuration when the proportion of high-bit quantization layers is increased to 30\%, corresponding to KVmix-k2.28v2.56. Table \ref{table012} supplements the performance data of various models under different KVmix quantization schemes. The results indicate that KVmix-k2.19v2.38 achieves a significant improvement in accuracy compared to KVmix-2bit and random-k2.19v2.38, highlighting the advantages of importance-aware quantization. When compared to KVmix-4bit, KVmix-k2.19v2.38 maintains an average accuracy loss within 1.30\% while achieving nearly twice the KV Cache compression ratio. Furthermore, increasing the proportion of high-bit quantization layers to 30\% (KVmix-k2.28v2.56) leads to an enhancement in model accuracy, approaching that of KVmix-4bit. These findings validate the flexibility of the KVmix framework in balancing accuracy and efficiency.

\begin{table*}[htbp]
\centering
\small
\scalebox{0.88}{
\setlength{\tabcolsep}{1.0mm}{
\begin{tabular}{l l c c c c c c c c c c c}
\toprule
\multirow{2}{*}{\textbf{Models}} & \multirow{2}{*}{\textbf{Methods}} & \multicolumn{10}{c}{\textbf{Datasets}} & \multirow{2}{*}{\textbf{Average}} \\
\cmidrule{3-12}
& & \textbf{\rotatebox{50}{TriviaQA}} & \textbf{\rotatebox{50}{Qasper}} & \textbf{\rotatebox{50}{MF-en}} & \textbf{\rotatebox{50}{QMSum}} & \textbf{\rotatebox{50}{2WikiMQA}} & \textbf{\rotatebox{50}{GovRep}} & \textbf{\rotatebox{50}{Rbench-P}} & \textbf{\rotatebox{50}{LCC}} & \textbf{\rotatebox{50}{TREC}} & \textbf{\rotatebox{50}{PsgRetr-en}} & \\
\midrule
\multirow{6}{*}{Llama-2-7B-hf} 
    & FP16  & 78.89 & 9.55  & 22.86 & 21.19 & 9.94  & 17.36 & 55.64 & 66.70 & 66.00 & 6.64  & 35.477 \\
    & KVmix-4bit & 77.88 & 9.11  & 22.43 & 21.13 & 9.75  & 17.53 & 56.17 & 66.70 & 66.00 & 6.35  & 35.305 \\
    & KVmix-2bit & 77.57 & 9.58  & 22.47 & 20.45 & 9.15  & 14.03 & 56.34 & 66.18 & 66.00 & 5.29  & 34.706 \\
    & random-k2.19v2.38& 78.30 & 9.39  & 22.54 & 20.41 & 9.46  & 14.01 & 56.36 & 66.31 & 66.00 & 5.49  & 34.827 \\
    %%\rowcolor{gray!20}
    & \cellcolor{gray!20}\textbf{KVmix-k2.19v2.38} & \cellcolor{gray!20}\textbf{78.78} & \cellcolor{gray!20}\textbf{9.59}  & \cellcolor{gray!20}\textbf{22.82} & \cellcolor{gray!20}\textbf{20.49} & \cellcolor{gray!20}\textbf{9.77}  & \cellcolor{gray!20}\textbf{15.45} & \cellcolor{gray!20}\textbf{56.54}  & \cellcolor{gray!20}\textbf{66.59} & \cellcolor{gray!20}\textbf{66.00}  & \cellcolor{gray!20}\textbf{5.72}  & \cellcolor{gray!20}\textbf{35.175} \\
    %%\rowcolor{gray!20}
    & \cellcolor{gray!20}\textbf{KVmix-k2.28v2.56} & \cellcolor{gray!20}\textbf{78.05} & \cellcolor{gray!20}\textbf{10.21} & \cellcolor{gray!20}\textbf{23.21} & \cellcolor{gray!20}\textbf{20.63}  & \cellcolor{gray!20}\textbf{9.72} & \cellcolor{gray!20}\textbf{15.62} & \cellcolor{gray!20}\textbf{56.61} & \cellcolor{gray!20}\textbf{66.64} & \cellcolor{gray!20}\textbf{66.00} & \cellcolor{gray!20}\textbf{6.08}  & \cellcolor{gray!20}\textbf{35.277} \\
\midrule
\multirow{5}{*}{Llama-3-8B-Instruct} 
    & FP16       & 78.35 & 40.75 & 46.80 & 21.69 & 32.39 & 30.62 & 49.77 & 56.51 & 70.50 & 37.00 & 46.438 \\
    & KVmix-4bit & 78.30 & 41.39 & 46.31 & 21.58 & 32.80 & 30.31 & 49.21 & 56.73 & 70.50 & 37.00 & 46.413 \\
    & KVmix-2bit & 76.13 & 39.18 & 45.70 & 21.20 & 32.19 & 29.62 & 44.56 & 49.10 & 71.00 & 36.30 & 44.498 \\
    & random-k2.19v2.38 & 78.01	& 39.17	& 45.90	& 21.22	& 32.02	& 29.55	& 45.36	& 49.71	& 71.00 & 36.50 & 44.844 \\
    %%\rowcolor{gray!20}
    & \cellcolor{gray!20}\textbf{KVmix-k2.19v2.38} & \cellcolor{gray!20}\textbf{78.13} & \cellcolor{gray!20}\textbf{39.15} & \cellcolor{gray!20}\textbf{46.31} & \cellcolor{gray!20}\textbf{21.26} & \cellcolor{gray!20}\textbf{32.20} & \cellcolor{gray!20}\textbf{29.95} & \cellcolor{gray!20}\textbf{47.56} & \cellcolor{gray!20}\textbf{51.62} & \cellcolor{gray!20}\textbf{71.00} & \cellcolor{gray!20}\textbf{36.50} & \cellcolor{gray!20}\textbf{45.368} \\
    %%\rowcolor{gray!20}
    & \cellcolor{gray!20}\textbf{KVmix-k2.28v2.56} & \cellcolor{gray!20}\textbf{78.27} & \cellcolor{gray!20}\textbf{39.30} & \cellcolor{gray!20}\textbf{46.42} & \cellcolor{gray!20}\textbf{21.11} & \cellcolor{gray!20}\textbf{32.82} & \cellcolor{gray!20}\textbf{29.92} & \cellcolor{gray!20}\textbf{49.13} & \cellcolor{gray!20}\textbf{53.92} & \cellcolor{gray!20}\textbf{71.00} & \cellcolor{gray!20}\textbf{37.00} & \cellcolor{gray!20}\textbf{45.889} \\
    % & random-k2.19v2.38& 76.13 & 39.18 & 45.7  & 21.2  & 32.19 & 29.62 & 44.56 & 71   & 36.3  & 44.498 \\
    % \rowcolor{gray!20}
    % & \textbf{KVmix-k2.19v2.38} & \textbf{78.13} & \textbf{39.15} & \textbf{46.31} & \textbf{21.26} & \textbf{32.2}  & \textbf{29.95} & \textbf{47.56} & \textbf{71}   & \textbf{36.5}  & \textbf{45.868} \\
    % & KVmix-k2.28v2.56& 78.27 & 39.3  & 46.42 & 21.11 & 32.82 & 29.32 & 49.13 & 67.5 & 37.5  & 41.016 \\
\midrule
\multirow{6}{*}{Llama-3.1-8B} 
    & FP16 & 83.67 & 11.53 & 31.13 & 22.88 & 13.92 & 29.23 & 61.84 & 68.96 & 67.50 & 19.50 & 41.016 \\
    & KVmix-4bit & 83.77 & 11.75 & 31.44 & 22.59 & 13.85 & 29.04 & 61.20 & 69.28 & 67.50 & 19.00 & 40.942 \\
    & KVmix-2bit & 83.10 & 10.90 & 30.76 & 22.11 & 13.08 & 27.02 & 58.92 & 68.13 & 67.00 & 19.00 & 40.002 \\
    & random-k2.19v2.38 & 83.25 & 10.90 & 31.05 & 22.34 & 13.05 & 27.01 & 59.26 & 68.39 & 67.00 & 19.00 & 40.125 \\
    %%\rowcolor{gray!20}
    & \cellcolor{gray!20}\textbf{KVmix-k2.19v2.38} & \cellcolor{gray!20}\textbf{83.28} & \cellcolor{gray!20}\textbf{11.40} & \cellcolor{gray!20}\textbf{31.49} & \cellcolor{gray!20}\textbf{22.90} & \cellcolor{gray!20}\textbf{12.92} & \cellcolor{gray!20}\textbf{27.22} & \cellcolor{gray!20}\textbf{59.96} & \cellcolor{gray!20}\textbf{68.54} & \cellcolor{gray!20}\textbf{67.50} & \cellcolor{gray!20}\textbf{19.50} & \cellcolor{gray!20}\textbf{40.471} \\
    %%\rowcolor{gray!20}
    & \cellcolor{gray!20}\textbf{KVmix-k2.28v2.56} & \cellcolor{gray!20}\textbf{83.80} & \cellcolor{gray!20}\textbf{11.37} & \cellcolor{gray!20}\textbf{31.31} & \cellcolor{gray!20}\textbf{22.85} & \cellcolor{gray!20}\textbf{13.47} & \cellcolor{gray!20}\textbf{27.75} & \cellcolor{gray!20}\textbf{60.94} & \cellcolor{gray!20}\textbf{68.10} & \cellcolor{gray!20}\textbf{67.50} & \cellcolor{gray!20}\textbf{19.50} & \cellcolor{gray!20}\textbf{40.659} \\

    % & FP16             & 83.67 & 11.53 & 31.13 & 22.88 & 13.92 & 27.84 & 61.84 & 67.5 & 19.5  & 40.942 \\
    % & KVmix-4bit      & 83.77 & 11.75 & 31.44 & 22.59 & 13.85 & 29.04 & 61.2  & 67.5 & 19    & 40.942 \\
    % & random-k2.19v2.38& 83.25 & 10.9  & 31.05 & 22.34 & 13.05 & 27.01 & 59.26 & 67   & 19    & 40.125 \\
    % \rowcolor{gray!20}
    % & \textbf{KVmix-k2.19v2.38} & \textbf{83.28} & \textbf{11.4} & \textbf{31.49} & \textbf{22.9} & \textbf{12.92} & \textbf{27.22} & \textbf{59.96} & \textbf{67.5} & \textbf{19.5}  & \textbf{40.471} \\
    % & KVmix-k2.28v2.56& 83.8  & 11.37 & 31.31 & 22.85 & 13.47 & 27.75 & 60.94 & 67.5 & 19.5  & 40.659 \\
\midrule
\multirow{6}{*}{Mistral-7B-Instruct-v0.3} 
    & FP16 & 84.29 & 36.19 & 54.70 & 21.79 & 35.08 & 32.84 & 53.06 & 57.56 & 73.50 & 32.50 & 48.151 \\
    & KVmix-4bit & 84.49 & 35.83 & 54.35 & 22.39 & 35.08 & 32.83 & 52.69 & 57.99 & 73.50 & 32.50 & 48.165 \\
    & KVmix-2bit & 84.08 & 34.29 & 53.87 & 21.37 & 33.39 & 32.05 & 50.99 & 56.18 & 73.50 & 32.00 & 47.172 \\
    & random-k2.19v2.38 & 84.01 & 34.35 & 53.61 & 21.45 & 33.40 & 32.20 & 50.59 & 56.53 & 73.50 & 32.50 & 47.214 \\
    %%\rowcolor{gray!20}
    & \cellcolor{gray!20}\textbf{KVmix-k2.19v2.38} & \cellcolor{gray!20}\textbf{84.03} & \cellcolor{gray!20}\textbf{35.67} & \cellcolor{gray!20}\textbf{53.68} & \cellcolor{gray!20}\textbf{21.84} & \cellcolor{gray!20}\textbf{33.81} & \cellcolor{gray!20}\textbf{32.19} & \cellcolor{gray!20}\textbf{51.98} & \cellcolor{gray!20}\textbf{56.88} & \cellcolor{gray!20}\textbf{73.50} & \cellcolor{gray!20}\textbf{32.75} & \cellcolor{gray!20}\textbf{47.633} \\
    %%\rowcolor{gray!20}
    & \cellcolor{gray!20}\textbf{KVmix-k2.28v2.56} & \cellcolor{gray!20}\textbf{84.45} & \cellcolor{gray!20}\textbf{35.87} & \cellcolor{gray!20}\textbf{54.35} & \cellcolor{gray!20}\textbf{21.97} & \cellcolor{gray!20}\textbf{34.23} & \cellcolor{gray!20}\textbf{32.51} & \cellcolor{gray!20}\textbf{51.85} & \cellcolor{gray!20}\textbf{57.25} & \cellcolor{gray!20}\textbf{73.50} & \cellcolor{gray!20}\textbf{33.00} & \cellcolor{gray!20}\textbf{47.898} \\

    % & FP16             & 84.29 & 36.19 & 54.7  & 21.79 & 35.08 & 32.84 & 53.06 & 73.5 & 32.5  & 48.151 \\
    % & KVminx-4bit      & 84.49 & 35.83 & 54.35 & 21.39 & 35.08 & 32.63 & 52.69 & 73.5 & 32.5  & 48.165 \\
    % & random-k2.19v2.38& 84.01 & 34.35 & 53.61 & 21.45 & 33.4  & 32.2  & 50.59 & 73.5 & 32.5  & 47.214 \\
    % \rowcolor{gray!20}
    % & \textbf{KVmix-k2.19v2.38} & \textbf{84.03} & \textbf{35.67} & \textbf{53.68} & \textbf{21.84} & \textbf{33.81} & \textbf{32.19} & \textbf{51.98} & \textbf{73.5} & \textbf{32.75} & \textbf{47.633} \\
    % & KVmix-k2.28v2.56& 84.45 & 35.87 & 54.35 & 21.97 & 34.23 & 32.51 & 51.85 & 73.5 & 33    & 47.698 \\
\bottomrule
\end{tabular}}
}
\caption{Model accuracy of 4 LLMs on LongBench with different quantization configurations. KVmix-k2.28v2.56 uses the configurations of Fig. \ref{fig14}. KVmix-2bit uses the asymmetric 2-bit (Key per-channel and Value per-token) quantization for all model layers (RPC ratio is set to 10\%). KVmix-4bit uses the asymmetric 4-bit quantization for all model layers (RPC ratio is set to 20\%). random-k2.19v2.38 randomly selects 20\% of the model layers to perform asymmetric 3-bit and 4-bit quantization for Key and Value (RPC ratio is set to 20\%), and the remaining layers are 2-bit quantization (RPC ratio is set to 10\%).}
\label{table012}
\end{table*}

\end{document}